\pdfoutput=1
\documentclass{article}

\usepackage[main,final]{neurips_2026}

\usepackage[utf8]{inputenc}
\usepackage[T1]{fontenc}
\usepackage{hyperref}
\usepackage{url}
\usepackage{booktabs}
\usepackage{amsfonts}
\usepackage{nicefrac}
\usepackage{microtype}
\usepackage{xcolor}
\usepackage{float}

\usepackage{graphicx}
\usepackage{wrapfig}
\usepackage{colortbl}
\usepackage{makecell}
\usepackage{enumitem}
\usepackage{amsmath}
\usepackage{amssymb}
\usepackage{mathtools}
\usepackage{amsthm}
\usepackage[capitalize,noabbrev]{cleveref}

\theoremstyle{plain}
\newtheorem{theorem}{Theorem}[section]

\newtheorem{lemma}[theorem]{Lemma}

\theoremstyle{definition}
\newtheorem{definition}[theorem]{Definition}
\newtheorem{assumption}[theorem]{Assumption}
\theoremstyle{remark}

\title{CONSTRAINER: Promptable Graph-Structured Optimization via Constraint Conditioning}

\author{%
  Zeeshan Memon$^{1}$ \quad
  Mingke Tian$^{1}$ \quad
  Xinyuan Song$^{1}$ \quad
  Hongwei Jin$^{2}$ \\
  \textbf{Kibaek Kim}$^{2}$ \quad
  \textbf{Liang Zhao}$^{1,3}$\thanks{Corresponding author: \texttt{liang.zhao@emory.edu}.} \\
  $^{1}$Emory University \quad
  $^{2}$Argonne National Laboratory \quad
  $^{3}$Causal Dynamics Lab \\
  \texttt{\{zeeshan.memon, liang.zhao\}@emory.edu}
}

\begin{document}

\maketitle

\begin{abstract}
Neural methods for graph-structured optimization have achieved strong performance within individual problem classes, but are typically trained for fixed formulations, limiting reuse across related problem variants. Recent progress in foundation models has motivated shared neural backbones trained across multiple graph-structured optimization tasks; however, existing approaches primarily share instance-level representations while treating feasibility constraints implicitly. A key source of shared structure across constrained optimization problems is the constraint set itself, which often exhibits substantial overlap across tasks.
We propose CONSTRAINER, a constraint-conditioned framework for graph-structured optimization that treats constraints as structured inputs. Constraint specifications are parsed into operator graphs and encoded using a graph neural network to produce invariant constraint embeddings, which condition a shared backbone model for solution prediction. We evaluate our approach on power grid optimization and mixed-integer linear programming problems, where it shows improved solution optimality and enables efficient transfer learning and generalization compared to baseline methods.
\end{abstract}

\section{Introduction}
\label{sec:intro}
Graph-structured optimization problems arise whenever decisions are made over interconnected entities whose interactions are naturally represented as nodes and edges, such as transportation and logistics networks, communication systems, supply chains, and power grids~\citep{li2025constraint, singh2022combinatorial}. In these settings, the graph encodes physical, informational, or economic dependencies, while optimization seeks to allocate resources, route flows, or coordinate actions across the network. As a result, graph-structured optimization problems are highly constrained by conservation laws, capacity limits, safety requirements, and inter-node coupling, and these constraints are often nonconvex, discrete, or temporally coupled, making large-scale instances challenging for classical numerical solvers. Despite the apparent stability of the underlying graph abstraction, real-world deployments are far from static. Any large-scale, safety-critical, graph-structured system that interacts with humans, markets, or uncertain environments inevitably requires frequent optimization reformulations, even when its graph representation and solver stack remain largely unchanged. Many operational settings impose strict latency requirements, under which rapid, near-optimal decisions are often preferable to exact solutions. While classical solvers are designed to provide strong feasibility and optimality guarantees, they can incur prohibitive computational cost as problem scale, model fidelity, or uncertainty increases. These considerations motivate growing interest in learning-based approaches to graph-structured optimization, where neural models are used to accelerate solving via warm-starting, heuristic learning~\citep{bengio2021machine}, or direct solution prediction~\citep{bello2016neural}, offering substantial efficiency gains when solving large numbers of related problem instances.

Recent progress in learning-based approaches has further accelerated this trend. Neural optimization models have begun to demonstrate generalization across distributions of problem instances within a fixed problem class, enabled by shared graph encoders, attention mechanisms, and large-scale pretraining~\citep{cai2025multi,yuanoptfm}. Parallel lines of work have explored diffusion models for generating distributions over near-optimal solutions~\citep{sun2023difusco}, as well as large language models used either as end-to-end solvers~\citep{jiang2025large} or as code generators conditioned on structured problem embeddings~\citep{li2025strcmp}. Similar ideas have also emerged in applied domains such as power-grid optimization, where foundation-style models aim to reuse a common backbone across related tasks~\citep{hamann2024foundation}.
Despite this progress, existing foundation models for optimization primarily share problem representations, either at the backbone level or through instance-level encoders~\citep{yuanoptfm}, while treating feasibility constraints implicitly through task-specific architectural choices~\citep{zong2025unico}. As a result, adapting to a new task often still requires dedicated finetuning, architectural modifications, or problem-specific engineering, limiting systematic transfer across tasks.

In practice, this limitation arises because many reformulations are driven by modifications to the constraint set, including the introduction, removal, or tightening of constraints, reflecting evolving operating conditions, regulatory requirements, or reliability criteria, while decision variables and objectives are preserved. Although these reformulated problems may differ syntactically, their feasible regions are typically governed by a small number of recurring constraint families~\citep{regin2010global, leo2022globalizing, mattenet2021generic}, such as conservation laws and capacity limits in physical networks~\citep{zamzam2020learning, fioretto2018distributed}, or assignment, flow, and scheduling constraints in combinatorial settings~\citep{goal2025}. Hence, related optimization tasks can often be viewed as different compositions or parameterizations of shared constraint primitives rather than entirely independent problem classes. Failing to exploit this structural commonality treats each reformulation in isolation, whereas explicitly encoding and conditioning on constraint structure provides a principled basis for improved generalization and systematic reuse of learned representations~\citep{jaegle2021perceiver, zhu2025generalizing}.

Building on this perspective, our goal is to develop a foundation model for graph-structured optimization, a single, reusable model that can adapt to new problem reformulations by taking constraint specifications as prompts and producing high-quality solution predictions usable directly as candidates, as warm-starts for downstream solvers, or for large-scenario screening across many related problem instances. Achieving this requires addressing three core challenges. First, constraints in graph-structured optimization are inherently relational, encoding structured dependencies among variables and nodes; naive vectorization discards this structure and limits transfer~\citep{vastl2024symformer, ranasinghe2025graph}. Second, constraints frequently admit multiple equivalent formulations, such as algebraic reparameterizations of physical laws or alternative expressions of flow balance, introducing invariances that a general-purpose model must recognize to avoid overfitting to surface syntax. Third, effective adaptation across evolving constraints demands a principled mechanism for interleaving constraint representations with objectives and graph structure, while ensuring that the resulting model remains adaptable to new constraints and generalizable beyond those seen during training.
To address these challenges, we treat constraints as first-class structured inputs and condition learning explicitly on their representations. Constraint specifications are parsed into operator graphs that preserve compositional semantics and encoded into invariant embeddings using graph neural networks, inspired by prior work on symbolic regression and equation modeling~\citep{feng2021graphmr, xiang2025graph, chen2020compositional}. These embeddings modulate a shared optimization backbone, factorizing learning into a shared component capturing general optimization behavior and a constraint-specific component adapting the model to feasibility, enabling reuse across reformulations. Beyond empirical motivation, we provide a theoretical analysis establishing stability of the learned constraint embeddings under admissible reparameterizations, showing that small structural perturbations in constraint representations induce bounded changes in the embedding space. This result formalizes the invariance properties required for constraint-conditioned generalization and provides a principled foundation for adapting to new constraints without retraining.
Our \textbf{core contributions} are:
\begin{itemize}[leftmargin=4pt, topsep=0pt, itemsep=0pt, parsep=0pt]
\setlength{\itemsep}{0pt}
\setlength{\parsep}{0pt}
\setlength{\topsep}{0pt}
\item \textbf{Constraint-conditioned foundation model.}
We propose a constraint-conditioned learning framework for graph-structured optimization, introducing a structured constraint encoder that parses constraint specifications into operator graphs and learns invariant constraint embeddings.
\item \textbf{Structured conditioning for generalization and composition.}
We design a constraint-conditioned backbone that enables generalization across constraint families and supports compositional transfer across reformulated optimization tasks with shared objectives and variables.
\item \textbf{Theoretical guarantees under reparameterization.}
We provide a theoretical analysis establishing stability of learned constraint embeddings under admissible constraint reparameterizations, formally supporting adaptability and within-family generalization.
\item \textbf{Empirical validation across optimization domains.}
We demonstrate the effectiveness of our approach on mixed-integer linear programming (MILP) and power-grid optimization benchmarks, achieving improved generalization and more efficient transfer compared to existing baselines.
\end{itemize}

\section{Previous Work}
\label{sec:prev_work}
\paragraph{Neural optimization methods.}Learning-based solvers for optimization problems span neural combinatorial optimization (NCO), constrained generative modeling, and differentiable optimization. In NCO, models learn constructive policies that sequentially build solutions, from sequence decoders \citep{vinyals2015pointer} to attention- and graph-based architectures for routing and related problems \citep{kool2019attention,kwon2020pomo}. While effective in-distribution, these methods struggle under distribution and scale shift \citep{joshi2022rethinking}. Recent work improves robustness via instance-conditioned adaptation and architectural inductive biases \citep{zhou2024icam,fang2024invit,zheng2024udc}, combined with test-time refinement or projection to reduce optimality gaps and enforce feasibility \citep{li2023t2t,li2024fastt2t,construct2024,sal2025}. Parallel efforts address feasibility through relational reasoning over variable--constraint graphs \citep{selsam2018singlesat,gasse2019exact}, relaxation- and meta-learning-based objectives \citep{wang2022objective,wang2023metalearning}, and constrained generative models, including diffusion-based solvers \citep{engel2017latentconstraints,dhariwal2021guidance,sun2023difusco,sanokowski2024diffusionunsup}. For continuous problems such as grid optimization, learning-based approaches predict primal solutions or warm-start classical solvers, such as in AC optimal power flow \citep{fioretto2020predicting,zamzam2020learning}, or embed differentiable optimization layers to impose hard constraints via implicit differentiation \citep{amos2017optnet, agrawal2019differentiable}. Across these settings, constraint semantics are enforced implicitly through task- or parameterization-specificity or learned from supervised losses. Differentiable repair layers, such as DC3~\citep{donti2021dc3}, FSNet~\citep{nguyen2025fsnet}, and bisection projection~\citep{liang2025bisection}, embed feasibility enforcement within a fixed formulation through a projection step composed with the predictor.
\paragraph{Foundation models for optimization.} Recent work explores optimization agents that share backbones across multiple tasks~\citep{goal2025}. Other efforts pursue unification through problem reductions or common interfaces \citep{pan2025unico, zong2025unico}, large-scale graph pretraining with multi-view objectives \citep{yuanoptfm}, and scalable decoding architectures such as parallel autoregressive modeling \citep{berto2025parco}.
Parallel work integrates language-model priors with structural inductive biases to improve generalization and controllability \citep{li2025strcmp}.
Beyond end-to-end solvers, learned solver selection and algorithm configuration provide decision layers on optimization pipelines \citep{gao2025solverselection,reijnen2025graph}, and standardized benchmarks aim to unify evaluation in classic graph problems \citep{ma2025ml4co}. Similarly, grid optimization problems have also been seen through the lens of foundation models and utilize a common backbone for different grid tasks~\citep{hamann2024foundation, puech2024optimal}.
Collectively, these works demonstrate the potential of co-training and shared representations to exploit structural commonalities across optimization problems, but they largely share instance- or problem-level representations rather than constraint semantics.
\paragraph{Learning-based MILP solution prediction.} Predict-and-search and related methods predict, or differentiate through, solutions of a single fixed MILP instance~\citep{han2023gnn,huang2024contrastive,liu2025apollo,geng2025differentiable}, encoding it as a bipartite variable--constraint graph in which the constraints are consumed numerically. CONSTRAINER encodes the algebraic form of the constraint set as a promptable input, so a related or reformulated problem becomes a new input to the same model. The two directions are complementary: CONSTRAINER's predictions can serve as the prediction stage of such pipelines.
\section{Methodology}
\label{sec:method}
\subsection{Problem Description}
Graph-structured optimization problems over discrete and continuous domains can be described within a unified framework, in which decision variables are associated with the nodes and edges of a graph and feasibility is determined by an explicit set of constraints, while an objective function specifies optimality over the feasible set. 
A graph-structured constrained optimization can be represented as:
\begin{equation}
\mathcal{P} = (\mathcal{G}, \mathcal{C}, f),
\end{equation}
\begin{wrapfigure}{r}{0.55\linewidth}
    \centering
    \includegraphics[width=\linewidth]{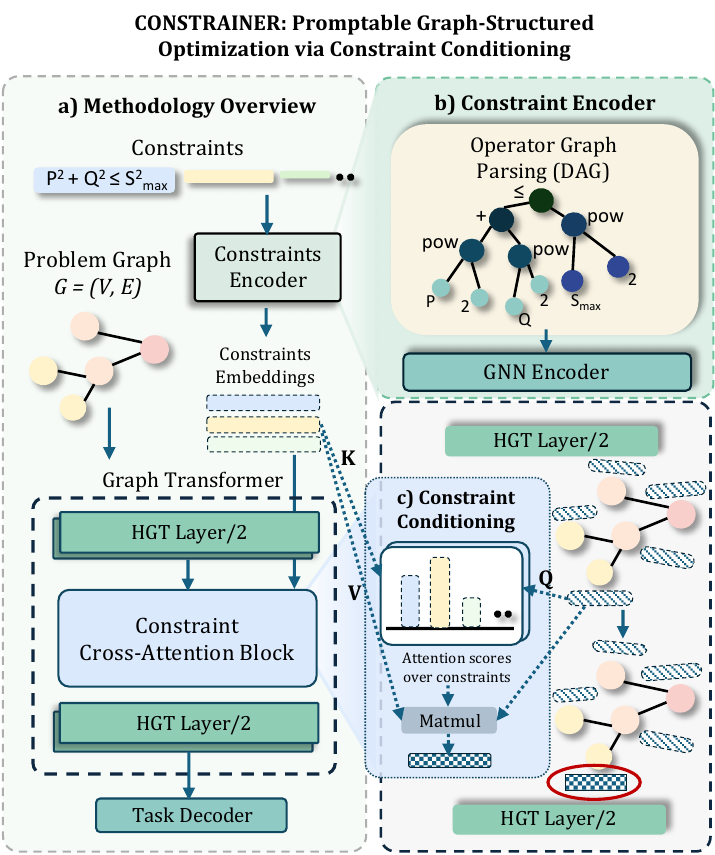}
    \caption{a) Methodology Overview. b) Constraint encoder parses each constraint into a DAG and then encodes it with a GNN. c) Constraint-conditioning block: each node attends to all constraints.}
    \label{fig:method}
\end{wrapfigure}
where \( \mathcal{G} = (V, E) \) describes the structural organization of the problem, \( \mathcal{C} \) is a set of constraints, and \( f \) is an objective function.
The graph \( \mathcal{G} \) specifies the entities involved and their relationships, such as cities connected by distances in routing problems or buses connected by transmission lines in power grid optimization. Decision variables are associated with nodes or edges of the graph, and the graph structure specifies which variables interact. Differences between combinatorial and continuous optimization problems arise from the domains and constraints of the decision variables, while the graph abstraction provides a common representation for organizing variables and their interactions.

Let \( \Omega \) denote the space of all possible value assignments to the decision variables associated with the nodes and edges of \( \mathcal{G} \). Each element \( x \in \Omega \) represents a complete assignment of values to all decision variables. The constraint set \( \mathcal{C} \) consists of explicit mathematical relations over these variables, including equality and inequality constraints as well as domain restrictions such as integrality or discreteness. Each constraint $c \in \mathcal{C}$ is a predicate over assignments, and $c(x) = \text{true}$ indicates that $x$ satisfies $c$.
The feasible set induced by the problem \( \mathcal{P} \) is
\begin{equation}
\mathcal{X} = \left\{ x \in \Omega \;\middle|\; \forall c \in \mathcal{C},  c(x) = \text{true} \right\}.
\end{equation}
The objective function $f : \mathcal{X} \rightarrow \mathbb{R}$ assigns a cost to each feasible assignment, and for a given problem instance the optimal solution is $x^\star \in \arg\min_{x \in \mathcal{X}} f(x)$.
Our goal is to learn a constraint-conditioned foundation model $F_\theta$ that takes the graph structure $\mathcal{G}$ and constraint specification $\mathcal{C}$ as inputs and produces a high-quality solution,
\begin{equation}
\hat{x} = F_\theta(\mathcal{G}, \mathcal{C}),
\end{equation}
such that $\hat{x}$ satisfies the constraints in $\mathcal{C}$ and achieves optimal objective cost.
\\
\textbf{Motivation.}
As the constraint set \( \mathcal{C} \) determines feasibility and often recurs across tasks and instances, we explicitly condition a shared backbone neural model on representations of the constraint set to enable generalization across different problems and tasks by reusing constraint semantics, rather than representations of separate learning modules for each task.

\subsection{Encoding Constraints}

Constraints define the feasible set $\mathcal{X}$ of a problem instance
$\mathcal{P} = (\mathcal{G}, \mathcal{C}, f)$ and specify structured relations
among decision variables that determine feasibility. To enable reuse of
constraint semantics across problem instances and tasks, we represent each
constraint $c \in \mathcal{C}$ as a structured object that captures its
algebraic form while remaining robust to equivalent algebraic transformations.
Each constraint is represented as a directed acyclic graph (DAG)
\begin{equation}\label{eq:DAG}
\mathcal{A}_c = (V_c, E_c),
\end{equation}
corresponding to the computation graph of the constraint expression. Nodes
$v \in V_c$ represent operators, decision variables, or numerical constants,
while directed edges encode functional dependencies between operands and
operators. The acyclic structure reflects the compositional nature of algebraic
constraints and preserves operator hierarchy, argument roles, and evaluation
direction. This operator-graph representation captures each constraint at the level of its algebraic form, supporting both linear and nonlinear relations under a uniform encoding and enabling contrastive training for invariance under admissible reformulations. Node features are initialized according to node type:
\begin{equation}\label{eq:hv}
h_v^{(0)} =
\begin{cases}
\phi_{\text{var}}(v), & v \text{ is a variable node}, \\
\phi_{\text{const}}(v), & v \text{ is a numerical constant or bound}, \\
\phi_{\text{op}}(v), & v \text{ is an operator node},
\end{cases}
\end{equation}
where $\phi_{\text{var}}$, $\phi_{\text{const}}$, and $\phi_{\text{op}}$ are learned lookup tables for variable types, numerical primitives, and operator types, respectively. Constraint encoding uses message passing over $\mathcal{A}_c$ for $L$ layers:
\begin{equation}\label{eq:hl}
h_v^{(\ell+1)} =
\psi^{(\ell)} \left(
h_v^{(\ell)},
\sum_{u : (u,v) \in E_c}
\phi^{(\ell)} \left(h_u^{(\ell)}, e_{uv}\right)
\right),
\end{equation}
for $ \ell = 0, \dots, L-1$, producing contextualized node representations that capture the compositional and
relational structure of the constraint. A permutation-invariant attention pooling operator aggregates the final node embeddings to
produce a fixed-dimensional constraint representation:
\begin{equation}\label{eq:zc_pool}
z_c =
\textsc{Pool} \left(\{ h_v^{(L)} \mid v \in V_c \}\right)
 \equiv 
\phi_\theta(c).
\end{equation}
This representation is invariant to node ordering and graph isomorphisms that
preserve constraint semantics. To encourage invariance to algebraic and syntactic equivalences expressible at the DAG level, we train
constraint encoders using contrastive learning. For each constraint $c$, a
positive set contains constraints related by admissible DAG-level transformations
(e.g., rescaling or operator-graph rearrangement), while negative samples correspond
to structurally similar but semantically distinct constraints that alter
feasibility. Given a positive constraint $c^+$ and negatives
$\{c_k^-\}_{k=1}^K$, the contrastive objective is
\begin{equation}\label{eq:loss}
\mathcal{L}_{\text{con}}(c) = -\log \frac{\exp \left(\text{sim}(z_c, z_{c^+}) / \tau\right)}{\exp \left(\text{sim}(z_c, z_{c^+}) / \tau\right) + \sum_{k=1}^K \exp \left(\text{sim}(z_c, z_{c_k^-}) / \tau\right)},
\end{equation}
where $\text{sim}(\cdot,\cdot)$ denotes cosine similarity and $\tau$ is a
temperature parameter. Details of positive and negative constraint construction and training are provided in Appendix~\ref{subsec:imple_details}. The class of constraints expressible in this framework is precisely those reducible to a finite computation DAG over the framework's operator vocabulary; we characterize the scope in Appendix~\ref{app:scope}.

\subsection{Constraint-Conditioned Backbone}

Given a problem instance $\mathcal{P} = (\mathcal{G}, \mathcal{C}, f)$, the model
computes shared representations of decision variables over the problem structure
$\mathcal{G} = (V, E)$ while conditioning on the encoded constraint set
$\mathcal{C}$. Each node $i \in V$ may be associated with a set of instance-level features (e.g., power, capacity, demand). Rather than encoding these features as unstructured numeric vectors, we represent each feature as a \emph{typed feature token} that is semantically aligned with the variable representations used in constraint encoding. Concretely, each feature token $\tau$ is defined by a feature type and a scalar value, and is embedded as
\begin{equation}
    \mathbf{h}_{\tau}
    =
    \phi_{\text{var}}(\tau)
    \oplus
    \phi \left(\text{value}(\tau)\right),
\end{equation}
where $\phi_{\text{var}}$ denotes the same learnable type embedding function used for decision variable nodes in the constraint encoder, applied here according to the feature type associated with $\tau$, $\phi(\cdot)$ encodes scalar feature values, and $\oplus$ denotes element-wise addition. This shared embedding space aligns instance-level features with abstract variable symbols appearing in constraints. The initial embedding of variable $i$ is obtained by aggregating all feature tokens associated with $i$ using a permutation-invariant attention pooling of the same form as in Eq.~\eqref{eq:zc_pool}. Each token is weighted by a learned attention score that depends on its feature type, so exchanging the values of two features changes $\mathbf{h}_i^{(0)}$ and the representation preserves the binding between a feature type and its value:
\begin{equation}
    \mathbf{h}_i^{(0)}
    =
    \textsc{Aggregate}
    \left(
        \left\{
            \mathbf{h}_{\tau}
             \middle| 
            \tau \text{ applies to } i
        \right\}
    \right).
\end{equation}
Let $h_i^{(0)} := \mathbf{h}_i^{(0)}$. The backbone is
implemented as a heterogeneous graph transformer, here denoted as \textsc{GT}, which propagates information over
$\mathcal{G}$ using attention-based message passing. The first $L/2$ layers
perform structural reasoning over the problem graph,
\begin{equation}
h_i^{(\ell+1)} =
\textsc{GT}^{(\ell)} \left(
h_i^{(\ell)}, \{ h_j^{(\ell)} : (j,i) \in E \}
\right),
\end{equation}
for $\ell = 0, \dots, \tfrac{L}{2}-1$, producing intermediate variable representations that encode interactions induced
by the problem structure. Edge information, when present, is incorporated
implicitly through the graph transformer’s attention mechanism, following formulations in~\cite{goal2025}.
From the constraint encoder, each constraint $c \in \mathcal{C}$ is represented
by an embedding $z_c \in \mathbb{R}^d$, and the full constraint set is represented as
\begin{equation}
Z_{\mathcal{C}} = \{ z_c \}_{c \in \mathcal{C}} \subset \mathbb{R}^{d}.
\end{equation}
Constraint conditioning is introduced via a cross-attention operation, where
variable representations act as queries and constraint embeddings serve as keys
and values. For each variable $i$, a constraint-conditioned context vector is
computed as
\begin{equation}
\tilde{h}_i =
\sum_{c \in \mathcal{C}}
\alpha_{ic}  W_v z_c,
\quad
\alpha_{ic} =
\textsc{Softmax}_c
\left(
\left\langle
W_q h_i^{(L/2)}, 
W_k z_c
\right\rangle
\right),
\end{equation}
with $W_q$, $W_k$, and $W_v$ denoting learnable projections. The resulting context
is incorporated into the variable representation through a residual update,
\begin{equation}
h_i^{(L/2)} \leftarrow h_i^{(L/2)} + \tilde{h}_i.
\end{equation}
The remaining $L/2$ graph transformer layers propagate the
constraint-conditioned information over the problem graph,
\begin{equation}
h_i^{(\ell+1)} =
\textsc{GT}^{(\ell)} \left(
h_i^{(\ell)}, \{ h_j^{(\ell)} : (j,i) \in E \}
\right),
\end{equation}
for $ \ell = \tfrac{L}{2}, \dots, L-1$, allowing constraint effects to diffuse across structurally related variables. The resulting representations encode both global problem structure and
constraint-conditioned context.

\subsection{Task Decoder}

Let $H = \{h_i^{(L)}\}_{i \in V}$ denote the set of variable representations produced by
the constraint-conditioned backbone. The task decoder implements a mapping
\begin{equation}
\hat{x} = D_\theta(H),
\end{equation}
which predicts a solution either in a single forward pass for continuous-variable
problems or through a sequence of masked prediction steps for discrete
decision problems. For problems with continuous decision variables, decoding is performed
independently and in parallel over variables using a readout network,
\begin{equation}
\hat{x}_i = g_\theta(h_i), \qquad i \in V,
\end{equation}
where $g_\theta$ is a lightweight MLP applied to each variable representation.
Although decoding factorizes over variables, inter-variable dependencies induced
by constraints and problem structure are captured in the backbone
representations through graph-based message passing and constraint conditioning.
The model is trained with supervised losses using optimal solutions, augmented with penalty terms that measure constraint violation.
The feasibility mechanism depends on the problem class. When the binding constraints can be maintained incrementally over a discrete action set, as in the combinatorial problems, an autoregressive decoder that masks infeasible actions at each step is feasible by construction, following GOAL~\citep{goal2025}. When feasibility is dominated by continuous power-flow and thermal constraints, as in the grid problems including the mixed-integer DCOTS and DCTNEP, it is handled by the constraint-violation penalty, optionally composed with a projection step (Table~\ref{tab:hybrid_fsnet}). Within each problem class, the same feasibility setup applies to CONSTRAINER and the baselines.
\section{Theoretical Analysis}
\subsection{Stability to Constraint Reparameterization}
\label{sec:4.1}
We analyze stability of the constraint embedding $z_c$ under admissible reparameterizations of the operator-DAG representation. An admissible reparameterization is a rewrite $T$ of a constraint $c$ that leaves its feasible set unchanged, $\mathcal{X}(c)=\mathcal{X}(T(c))$ after projecting out auxiliary variables such as slacks; examples include positive rescaling, operand reordering, equality splitting, slack introduction, and big-M versus indicator encodings. Under a linear instantiation of message passing with bounded parameters and bounded typed-adjacency operators, the embedding difference $\|z_c - z_{c'}\|_2$ with $c' = T(c)$ admits an explicit upper bound in terms of a single \emph{reparameterization budget}
\[
\delta := \max_{r\in\{1,\dots,R\}} \|A_{c,r} - A_{c',r}\|_2,
\]
where $A_{c,r}\in\{0,1\}^{n_c\times n_c}$ is the type-$r$ adjacency matrix of the operator DAG for $c$. The bound is monotone in $\delta$: structurally closer reparameterizations yield smaller embedding shifts.
\begin{assumption}[Linear constraint encoder with bounded message weights]
\label{assump:linear-encoder-boundedW}
For a constraint $c$ with operator DAG defined in Eq.~\eqref{eq:DAG}, let the encoder hidden states be $h_{v,c}^{(\ell)}\in\mathbb{R}^{d}$ (Eq.~\eqref{eq:hv}) and be stacked row-wise as $H_c^{(\ell)}\in\mathbb{R}^{n_c\times d}$. We assume the following linear instantiation:
\begin{equation}
h_v^{(0)}=W_{\mathrm{init}}x_v,\qquad W_{\mathrm{init}}\in\mathbb{R}^{d\times p}.
\end{equation}
\begin{equation}
m_{u\to v}^{(\ell)}=W_{r_{uv}}^{(\ell)}h_u^{(\ell)},\qquad W_r^{(\ell)}\in\mathbb{R}^{d\times d},\ \ell=0,\dots,L-1.
\end{equation}
Moreover, there exist constants $w_{\mathrm{init}}\ge 0$ and $w_{\ell,r}\ge 0$ such that $\|W_{\mathrm{init}}\|_2\le w_{\mathrm{init}}$ and $\|W_r^{(\ell)}\|_2\le w_{\ell,r}$.
\end{assumption}

\begin{lemma}[Bounded parameters and graph degree bounds]
\label{lem:bounded-rigid}
Assume Assumption~\ref{assump:linear-encoder-boundedW}. Assume further that the linear instantiation of Eq.~\eqref{eq:hl} is
\begin{equation}
\phi^{(\ell)}(h_{u,c}^{(\ell)},e_{uv})=W_{r_{uv}}^{(\ell)}h_{u,c}^{(\ell)}=:m_{u\to v,c}^{(\ell)},
\end{equation}
\begin{equation}
\bar m_{v,c}^{(\ell)}=\sum_{u:(u,v)\in E_c}m_{u\to v,c}^{(\ell)},
\end{equation}
\begin{equation}
\psi^{(\ell)}(h_{v,c}^{(\ell)},\bar m_{v,c}^{(\ell)})=A^{(\ell)}h_{v,c}^{(\ell)}+B^{(\ell)}\bar m_{v,c}^{(\ell)},
\end{equation}
so that the update reduces to
\begin{equation}
h_{v,c}^{(\ell+1)}=A^{(\ell)}h_{v,c}^{(\ell)}+B^{(\ell)}\sum_{u:(u,v)\in E_c}m_{u\to v,c}^{(\ell)}.
\end{equation}
For each $r\in\{1,\dots,R\}$ define the type-$r$ adjacency matrix $A_{c,r}\in\{0,1\}^{n_c\times n_c}$ by $(A_{c,r})_{u,v}=1$ if and only if $(u,v)\in E_c$ and $r_{uv}=r$. Then, for each $\ell$ there exist constants $a_\ell,b_\ell\ge 0$ and for each $r$ there exists $\Delta_{c,r}\ge 0$ such that
\begin{equation}
\|A^{(\ell)}\|_2\le a_\ell,\qquad \|B^{(\ell)}\|_2\le b_\ell,\qquad \|A_{c,r}\|_2\le \Delta_{c,r}.
\end{equation}
\end{lemma}

\begin{theorem}[One-line stability bound]
\label{thm:one-line-stability}
Assume Assumption~\ref{assump:linear-encoder-boundedW} and the linear update in Lemma~\ref{lem:bounded-rigid}. Further assume that the compared constraints $c$ and $c'$ have the same number of nodes and the same initialization, i.e., $n_c=n_{c'}=:n$ and $H_c^{(0)}=H_{c'}^{(0)}=:H^{(0)}$. Define $a:=\max_\ell a_\ell$, $b:=\max_\ell b_\ell$, $w:=\max_{\ell,r} w_{\ell,r}$, and $\Delta:=\max_{q\in\{c,c'\},\,r}\|A_{q,r}\|_2$. Let $\delta$ be the reparameterization budget defined above. Then their embeddings satisfy
\begin{equation}
\|z_c-z_{c'}\|_2\le \|S\|_2\,\|\mathrm{vec}(H^{(0)})\|_2\cdot L\cdot bwR\delta\cdot\left(a+bwR\Delta\right)^{L-1},
\end{equation}
where $S:=I_d\otimes\mathbf{1}_n^\top$. 
\end{theorem}

The proof is in Appendix~\ref{sec:proof3}; a discussion of Assumption~\ref{assump:linear-encoder-boundedW} as standard MPNN/perturbation-theory conditions appears in Appendix~\ref{app:assumption-discussion}. The bound combines four scaling factors: pooling magnitude $\|S\|_2 \sim \sqrt{n}$, initial-state magnitude $\|\mathrm{vec}(H^{(0)})\|_2$, a depth term governed by the per-layer gain $a + b w R \Delta$ (combining self-updates, message injection, message transforms, edge types, and typed-adjacency spectral size), and a $\delta$-linear term. When the per-layer gain is near one, the encoder is near-nonexpansive; larger gains amplify perturbations with depth. Admissible reparameterizations keep $\delta$ small, so embedding shifts remain bounded.

\textbf{Scope.} The budget $\delta$ is defined between typed-adjacency matrices under a shared node indexing, so Theorem~\ref{thm:one-line-stability} applies to reparameterizations that preserve the variable and operator sets. Positive rescaling, operand reordering, slack introduction, and term rearrangements reduce to this case exactly: the parser canonicalizes linear and affine constraints to standard form before building the DAG, so forms such as $x+y\le 1$ and $y\le 1-x$ yield the same operator DAG and $\delta=0$. Rewrites that change the operator set, such as equality splitting, big-M versus indicator encodings, and integer versus Boolean formulations, lie outside the deterministic bound and are handled by contrastive training over admissible reparameterizations (Eq.~\eqref{eq:loss}, Table~\ref{tab:contras}), with generalization characterized by Theorem~\ref{thm:explicit-final}.
The contrastive objective in Eq.~\eqref{eq:loss} explicitly contracts embeddings of admissibly reparameterized constraints, tightening the bound of Theorem~\ref{thm:one-line-stability} in practice. Beyond stability to individual rewrites, contrastive training over admissible reparameterizations also enables generalization to unseen members of a constraint family: a coverage-based analysis yields a generalization bound (Theorem~\ref{thm:explicit-final}, Appendix~\ref{sec:4.2}) on the deviation between predictions on unseen constraints and on their nearest trained members of the same family. Full definitions and proofs are provided in Appendix~\ref{sec:4.2}.

\section{Experiments}
\label{sec:experiments}
\textbf{Datasets used.}
For continuous optimization, we consider power grid problems, including
Power Flow (PF, in AC and DC variants: ACPF, DCPF), AC Optimal Power Flow (ACOPF), DC Optimal Power Flow (DCOPF),
DC Optimal Transmission Switching (DCOTS), and
DC Transmission Network Expansion Planning (DCTNEP).
Problem instances are generated using standard formulations implemented in
\texttt{PowerModels.jl} and related extensions~\citep{coffrin2018powermodels}.
Continuous nonlinear problems (PF, ACOPF) are solved using IPOPT~\citep{wachter2006implementation},
while mixed-integer linear and quadratic formulations (DCOPF, DCOTS, DCTNEP)
are solved using the HiGHS solver~\citep{huangfu2018parallelizing}.
\begin{table}[t]
\centering
\footnotesize
\setlength{\tabcolsep}{4pt}
\renewcommand{\arraystretch}{1.05}
\caption{Performance comparison on power-grid optimization problems (IEEE-118).}
\label{tab:grid_results}
\begin{tabular}{l|c|c|cc|cc|cc|cc}
\hline
& \multicolumn{2}{c|}{PF}
& \multicolumn{2}{c|}{ACOPF}
& \multicolumn{2}{c|}{DCOPF}
& \multicolumn{2}{c|}{DCOTS}
& \multicolumn{2}{c}{DCTNEP} \\
Method
& ACPF Feas $\uparrow$
& DCPF Feas $\uparrow$
& Gap $\downarrow$ & Feas $\uparrow$
& Gap $\downarrow$ & Feas $\uparrow$
& Gap $\downarrow$ & Feas $\uparrow$
& Gap $\downarrow$ & Feas $\uparrow$ \\
\hline
GNN (CANOS)
& 95.11
& 97.14
& 4.80 & 93.00
& 3.60 & 94.20
& 6.10 & 88.45
& 5.40 & 85.32 \\

GAT
& 96.42
& 97.86
& 4.10 & 94.35
& 3.05 & 95.65
& 5.30 & 89.96
& 4.60 & 86.71 \\
\hline
GT (MTL)
& 97.88
& 98.72
& 3.55 & 95.60
& 2.35 & 97.10
& 4.90 & 90.85
& 3.80 & 87.94 \\
\hline
\rowcolor{blue!8}
CONSTRAINER
& \textbf{99.03}
& \textbf{99.50}
& \textbf{3.01} & \textbf{96.10}
& \textbf{1.50} & \textbf{99.02}
& \textbf{4.50} & \textbf{91.22}
& \textbf{2.90} & \textbf{88.91} \\
\hline
\end{tabular}
\end{table}
For MILP, we use the datasets of GOAL~\citep{goal2025}, which include combinatorial optimization problems across routing, scheduling, packing, graph, and location categories. Eight problems are used for training and the other eight for finetuning to show generalization compared to baselines. We evaluate all MILP tasks on fixed-size instances (100 nodes), with JSSP($10\times10$) and UMSP($100\times20$). Detailed problem definitions and data features are explained in Appendix~\ref{subsec:datasets}.
\newline 
\textbf{Baselines. }For power grid optimization, we compare against single- and multi-task neural baselines. For single-task learning, we include CANOS~\citep{piloto2024canos} and a GAT-based model~\citep{velickovic2018graph} for OPF~\citep{trigui2025graph}. For multi-task learning, we developed a baseline to evaluate a graph transformer trained jointly across multiple tasks with shared backbone and task-specific input/output adapters. For MILP problems, we evaluate against the following neural baselines: AM \citep{kool2019attention}, MDAM \citep{xin2021multi}, POMO \citep{kwon2020pomo}, Sym-NCO \citep{kim2022sym}, BQ-NCO \citep{drakulic2023bq}, MVMoE \citep{zhou2024mvmoe}, MatNet \citep{kwon2021matrix}, S2V-DQN \citep{khalil2017learning}, RouteFinder \citep{berto2024routefinder}, COMPASS \citep{chalumeau2023combinatorial}, Gumbeldore \citep{pirnay2024self}, OPTFM~\citep{yuanoptfm}, MTL~\citep{cai2025multi}, and GOAL~\citep{goal2025}. All baselines except MVMoE, RouteFinder, MTL, OPTFM and GOAL are trained as a single model per task. Further details of baseline implementations are discussed in Appendix~\ref{subsec:imple_details}.

\subsection{Performance on Grid and MILP Tasks}

\begin{table}[t]
\centering
\footnotesize
\setlength{\tabcolsep}{3pt}
\renewcommand{\arraystretch}{1.05}
\begin{minipage}[t]{0.49\textwidth}
\centering
\caption{Routing problems (optimality gap, \%, lower is better). Bold: best among generalist models.}
\label{tab:milp_routing_split}
\begin{tabular}{l|cccc}
\hline
Method & ATSP & CVRP & CVRPTW & OP \\
\hline
AM             & --     & 7.17 & --     & 5.39 \\
MDAM           & --     & 4.87 & --     & 2.91 \\
Sym-NCO        & --     & 3.36 & --     & 2.06 \\
POMO           & --     & 1.23 & --     & --     \\
MVMoE/4E       & --     & 1.68 & 4.95 & --     \\
RouteFinder-TE & --     & 1.53 & 3.24 & --     \\
BQ-NCO         & 1.30 & 2.81 & --     & 0.25 \\
MatNet         & 0.95 & --     & --     & --     \\
\hline
OPTFM          & 0.97 & 3.23 & 3.90 & 0.47 \\
MTL            & 0.94 & 3.19 & 3.86 & 0.45 \\
GOAL           & 0.91 & 3.16 & 3.82 & 0.43 \\
\hline
\rowcolor{blue!8}
CONSTRAINER    & \textbf{0.50} & \textbf{2.98} & \textbf{3.01} & \textbf{0.27} \\
\hline
\end{tabular}
\end{minipage}\hfill
\begin{minipage}[t]{0.49\textwidth}
\centering
\caption{Packing, graph, and scheduling problems (optimality gap, \%, lower is better). Bold: best among generalist models.}
\label{tab:milp_other_split}
\begin{tabular}{l|cccc}
\hline
Method & KP & MVC & UMSP & JSSP \\
\hline
POMO        & 0.20 & --     & --     & --     \\
BQ-NCO      & 0.11 & --     & --     & --     \\
S2V-DQN     & --     & 0.99 & --     & --     \\
COMPASS     & --     & --     & --     & 4.73 \\
Gumbeldore  & --     & --     & --     & 3.20 \\
\hline
OPTFM       & 0.14 & 0.40 & 3.95 & 4.27 \\
MTL         & 0.13 & 0.38 & 3.89 & 4.18 \\
GOAL        & 0.12 & \textbf{0.37} & 3.84 & 4.13 \\
\hline
\rowcolor{blue!8}
CONSTRAINER & \textbf{0.11} & 0.38 & \textbf{2.14} & \textbf{3.68} \\
\hline
\end{tabular}
\end{minipage}
\end{table}
CONSTRAINER achieves lower optimality gaps than OPTFM, MTL, and GOAL on seven of the eight tasks, as shown in Tables~\ref{tab:milp_routing_split} and~\ref{tab:milp_other_split}. The performance gains are more pronounced on routing and scheduling tasks, where feasibility is determined by multiple interacting constraints rather than a single dominant constraint family. In contrast, improvements on simpler packing and graph problems are limited, indicating that explicit constraint conditioning is most beneficial when constraint interactions occur. On power-grid optimization problems in Table~\ref{tab:grid_results}, CONSTRAINER achieves the highest feasibility rates across both continuous and mixed-integer formulations while maintaining lower optimality gaps than graph-only baselines. The largest gap reduction is observed on DCTNEP, which involves discrete expansion decisions coupled with power flow constraints. When strict feasibility is required on ACOPF, CONSTRAINER can be composed with downstream feasibility-seeking methods such as FSNet's L-BFGS refinement step; the resulting hybrid pipeline is compared with FSNet on Case30 and Case118 in Appendix~\ref{app:hybrid} (Table~\ref{tab:hybrid_fsnet}).
\subsection{Task Generalization Results}
Table~\ref{tab:grid_mtl} evaluates cross-task generalization to DCOPF. CONSTRAINER is evaluated without any task-specific finetuning, whereas MTL baselines are evaluated after 50-step adapter finetuning on DCOPF for fair comparison. Across all grid sizes, CONSTRAINER achieves lower optimality gaps and higher feasibility in the zero-shot setting compared to baselines. This reinforces that constraint-conditioned representations learned from ACPF, DCPF, and ACOPF transfer to DCOPF to a meaningful extent, even under unseen constraint composition.
A parallel out-of-distribution-size evaluation on ATSP, where models trained on 100-node instances are tested on 500- and 1000-node instances, is provided in Appendix~\ref{app:atsp_scale_gen} (Table~\ref{tab:atsp_scale_gen}); CONSTRAINER achieves the lowest optimality gap at every size.
\begin{figure}[t]
    \centering
    \includegraphics[width=\linewidth]{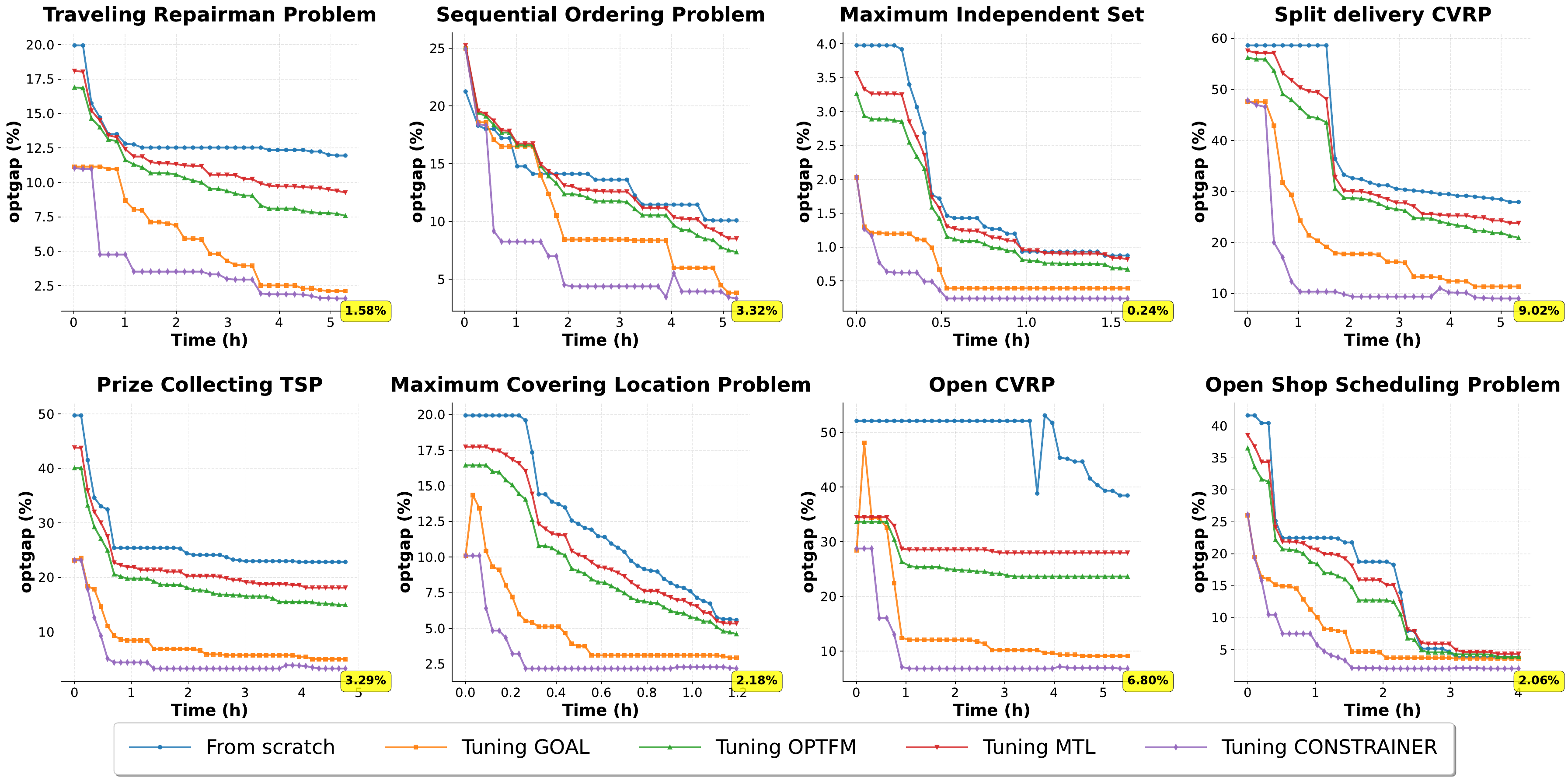}
    \caption{Finetuning CONSTRAINER on 8 unseen MILP tasks vs.\ baselines.}
    \label{fig:MILP_finetune}
\end{figure}
\begin{table}[t]
\centering
\small
\setlength{\tabcolsep}{2pt}
\renewcommand{\arraystretch}{1.05}
\begin{minipage}[t]{0.55\textwidth}
\centering
\caption{Generalization across grid sizes on DCOPF.}
\label{tab:grid_mtl}
\begin{tabular}{l|cc|cc|cc}
\hline
Method
& \multicolumn{2}{c|}{IEEE-30}
& \multicolumn{2}{c|}{IEEE-57}
& \multicolumn{2}{c}{IEEE-118} \\
 & Gap$\downarrow$ & Feas$\uparrow$
 & Gap$\downarrow$ & Feas$\uparrow$
 & Gap$\downarrow$ & Feas$\uparrow$ \\
\hline
\makecell[l]{MTL\\(DCPF adapter)}
& 7.11 & 93.11
& 11.33 & 92.93
& 6.35 & 91.11 \\
\makecell[l]{MTL\\(ACOPF adapter)}
& 5.88 & 92.00
& 9.11 & 90.31
& 6.50 & 90.01 \\
\rowcolor{blue!10}
\makecell[l]{\textbf{CONSTRAINER}}
& \textbf{1.58} & \textbf{98.90}
& \textbf{2.36} & \textbf{95.82}
& \textbf{3.13} & \textbf{94.65} \\
\hline
\end{tabular}
\end{minipage}\hfill
\begin{minipage}[t]{0.43\textwidth}
\centering
\caption{Ablation on constraint conditioning across routing tasks (optimality gap, \%).}
\label{tab:ablation_constraints}
\begin{tabular}{l|ccc}
\hline
Method & ATSP & CVRPTW & OP \\
\hline
\textbf{CONSTRAINER}
& \textbf{0.50} & \textbf{3.01} & \textbf{0.27} \\
\hline
\makecell[l]{Lang.\ Emb.\\constraints}
& 0.78 & 3.42 & 0.36 \\
\hline
\makecell[l]{No constraint\\cond.\ (MTL)}
& 0.94 & 3.86 & 0.45 \\
\hline
\end{tabular}
\end{minipage}
\end{table}
\newline
\textbf{Finetuning on unseen tasks.}
Across 8 unseen MILP problems, CONSTRAINER converges faster and stabilizes at lower optimality gaps than OPTFM, MTL, and GOAL, with the largest advantage in early finetuning (Fig.~\ref{fig:MILP_finetune}). On DCOTS and DCTNEP (Fig.~\ref{fig:DCOTS_finetune}), CONSTRAINER reaches convergence in approximately 30\% fewer epochs with lower final gaps. All combinatorial MILP tasks use a masked autoregressive decoder~\citep{goal2025} that excludes constraint-violating actions during decoding, so CONSTRAINER and the baselines all produce feasible outputs by construction. We therefore report optimality gap as the only varying metric on these problems.
\subsection{Ablation Study}
An ablation study is conducted to isolate the effect of constraint conditioning and to assess the role of structured constraint representations. Table~\ref{tab:ablation_constraints} compares CONSTRAINER against variants that progressively weaken or remove constraint conditioning across three routing problems.
Removing constraint conditioning entirely (multi-task baseline) consistently degrades performance across all tasks. Replacing the DAG encoder with language-based embeddings (text-embedding-3-small) partially recovers performance but stays below CONSTRAINER, indicating that surface-level semantic embeddings fail to preserve the compositional structure captured by the operator-graph encoding. An additional ablation on the layer at which constraint cross-attention is injected into the backbone is provided in Appendix~\ref{app:crossattn_placement} (Table~\ref{tab:crossattn_placement}); injecting at $L/2$ yields the best optimality gap on MIS. Appendix~\ref{app:review_analyses} reports significance tests, a matched-time comparison with CP-SAT, and the sensitivity of constraint embeddings to perturbations and reformulations.
\section{Conclusion}
\label{sec:conclusion}
We presented CONSTRAINER, a constraint-conditioned foundation model for graph-structured optimization that encodes constraint specifications as operator graphs and conditions a shared backbone on invariant constraint embeddings, enabling systematic reuse across reformulated problems. We established theoretical stability of the learned representations under admissible reparameterizations, supporting within-family generalization. Empirical results on mixed-integer linear programming and power-grid optimization tasks demonstrated improved solution quality, higher feasibility, and more effective zero-shot and finetuned task transfer compared to adapter-based multi-task learning. 
\begin{ack}
This work was supported in part by the National Science Foundation (NSF) under Grant Nos.\ 2432418, 2414115, 2403312, 2318831, and 2521631; the National Institutes of Health (NIH) under Grant Nos.\ R01AG089806 and R01CA297856; and a Sony Research Award. The content is solely the responsibility of the authors and does not necessarily represent the official views of the NSF, NIH, or other funding organizations.
\end{ack}

\bibliographystyle{plainnat}
\bibliography{example_paper}

\appendix

\section{Constraint families, contrastive training, and family coverage}
\label{sec:4.2}
We model admissible constraint reparameterizations as a transformation set $\mathcal{T}$ that preserves feasibility and constraint geometry, and define a constraint family as the orbit $\mathcal{F} = \{T(c_0) : T \in \mathcal{T}\} \subseteq \mathcal{C}$. This formalizes the fact that many syntactically distinct constraint expressions correspond to the same underlying feasible region. Our goal in this section is to characterize how contrastive training enables generalization within such constraint families.
The constraint encoder is trained using the contrastive objective in Eq.~\eqref{eq:loss}, where positive samples correspond to admissible reparameterizations $c^+ = T(c)$ and negatives correspond to non-equivalent constraints. Given a finite training subset $\mathcal{F}_{\mathrm{train}} \subseteq \mathcal{F}$, we measure within-family generalization via the coverage radius $R_\phi(\mathcal{F}; \mathcal{F}_{\mathrm{train}})$ in embedding space (Definition~\ref{def:coverage}). By explicitly pulling together embeddings of reparameterized constraints, the contrastive objective encourages unseen constraints $c \in \mathcal{F}$ to be mapped close to some representative in $\mathcal{F}_{\mathrm{train}}$. As a result, CONSTRAINER generalizes within constraint families, reducing reparameterization-induced distribution shift and enabling stable downstream prediction under unseen but semantically equivalent constraints.

\begin{definition}[Coverage radius in embedding space]
\label{def:coverage}
Fix a family $\mathcal{F}$ and a finite subset $\mathcal{F}_{\mathrm{train}}\subseteq\mathcal{F}$. For any $c\in\mathcal{F}$, define the nearest-neighbor embedding distance
\begin{equation}
\Delta_\phi(c;\mathcal{F}_{\mathrm{train}})=\min_{c'\in\mathcal{F}_{\mathrm{train}}}\|\phi_\theta(c)-\phi_\theta(c')\|_2.
\end{equation}
Define the coverage radius of $\mathcal{F}$ under $\mathcal{F}_{\mathrm{train}}$ as
\begin{equation}
R_\phi(\mathcal{F};\mathcal{F}_{\mathrm{train}})=\sup_{c\in\mathcal{F}}\Delta_\phi(c;\mathcal{F}_{\mathrm{train}}).
\end{equation}
\end{definition}

\begin{assumption}[Orbit coverage on training subset]
\label{assump:orbit-cover}
For every $c\in\mathcal{F}$, there exist $c'\in\mathcal{F}_{\mathrm{train}}$ and $T\in\mathcal{T}$ such that $c=T(c')$.
\end{assumption}

\begin{assumption}[Lipschitz predictor]
\label{assump:lipschitz}
There exists $L_F>0$ such that, for any fixed problem graph $\mathcal{G}$ and objective $f$, and for any two stacked constraint-embedding matrices $Z,Z'\in\mathbb{R}^{m\times d}$ with the same number and ordering of constraints,
\begin{equation}
\|F_\theta(\mathcal{G},f,Z)-F_\theta(\mathcal{G},f,Z')\|\le L_F\|Z-Z'\|_F.
\end{equation}
\end{assumption}

\begin{lemma}[Lower bound on positive similarity]
\label{lem:loss-to-sim}
Fix a constraint $c$ and a positive sample $c^+=T(c)$ for some $T\in\mathcal{T}$. Let $\{c_k^-\}_{k=1}^K$ be the negative samples associated with the pair $(c,c^+)$ in Eq.~\eqref{eq:loss}, and let $\tau>0$ be the temperature. Assume that there exists $\rho^-\in[-1,1)$ such that $\mathrm{sim}(z_c,z_{c_k^-})\ge \rho^-$ for every $k=1,\dots,K$. If $\mathcal{L}_{\mathrm{con}}(c)\le \varepsilon$, then
\begin{equation}
\mathrm{sim}(z_c,z_{c^+})\ge \rho^-+\tau\left(\log K-\log(e^\varepsilon-1)\right).
\end{equation}
\end{lemma}
The proof is provided in Section~\ref{sec:proof4}.

\begin{lemma}[Contrastive invariance coverage-radius bound]
\label{lem:coverage}
Fix a family $\mathcal{F}=\{T(c_0):T\in\mathcal{T}\}$ and $\mathcal{F}_{\mathrm{train}}\subseteq\mathcal{F}$. Assume Assumption~\ref{assump:orbit-cover}, and assume that the embeddings are $\ell_2$-normalized. Further assume that there exists a uniform $\rho^-\in[-1,1)$ such that $\mathrm{sim}(z_{c'},z_{c_k^-})\ge\rho^-$ for all training anchors $c'\in\mathcal{F}_{\mathrm{train}}$ and their negatives, and that for every $c=T(c')\in\mathcal{F}$ the corresponding positive-pair loss satisfies $\mathcal{L}_{\mathrm{con}}(c',T(c'))\le\varepsilon$. If $\varepsilon>0$ and $\rho^-+\tau\left(\log K-\log(e^\varepsilon-1)\right)\le 1$, then the coverage radius in Definition~\ref{def:coverage} satisfies
\begin{equation}
R_\phi(\mathcal{F};\mathcal{F}_{\mathrm{train}})\le \sqrt{2\left(1-\rho^--\tau\left(\log K-\log(e^\varepsilon-1)\right)\right)}.
\end{equation}
\end{lemma}
The proof is provided in Section~\ref{sec:proof5}.

\begin{lemma}[Constraint-family generalization controlled by coverage radius]
\label{lem:gen-by-coverage}
Suppose $\mathcal{C}_{\mathrm{test}}\subseteq\mathcal{F}$ and training provides $\mathcal{F}_{\mathrm{train}}\subseteq\mathcal{F}$. Let $\hat x_{\mathrm{test}}=F_\theta(\mathcal{G},f,Z_{\mathcal{C}_{\mathrm{test}}})$. Let $\hat x_{\mathrm{nn}}$ be the model output when each $c\in\mathcal{C}_{\mathrm{test}}$ is replaced by a nearest neighbor $c^{\mathrm{nn}}\in\mathcal{F}_{\mathrm{train}}$ in embedding space, giving $Z_{\mathcal{C}_{\mathrm{nn}}}$. Under Assumption~\ref{assump:lipschitz}, we have
\begin{equation}
\begin{aligned}
\|\hat x_{\mathrm{test}}-\hat x_{\mathrm{nn}}\|
&\le L_F\|Z_{\mathcal{C}_{\mathrm{test}}}-Z_{\mathcal{C}_{\mathrm{nn}}}\|_F \\
&\le L_F\sqrt{|\mathcal{C}_{\mathrm{test}}|} R_\phi(\mathcal{F};\mathcal{F}_{\mathrm{train}}).
\end{aligned}
\end{equation}
\end{lemma}
The proof is provided in Section~\ref{sec:proof6}.

\begin{theorem}[Explicit bound]
\label{thm:explicit-final}
Assume the conditions of Lemmas~\ref{lem:coverage} and~\ref{lem:gen-by-coverage}. Then
\begin{equation}
\begin{aligned}
&\|F_\theta(\mathcal{G},f,Z_{\mathcal{C}_{\mathrm{test}}})-F_\theta(\mathcal{G},f,Z_{\mathcal{C}_{\mathrm{nn}}})\| \\
&\le L_F\sqrt{|\mathcal{C}_{\mathrm{test}}|}\sqrt{2\left(1-\rho^--\tau\left(\log K-\log(e^\varepsilon-1)\right)\right)}.
\end{aligned}
\end{equation}
\end{theorem}

Theorem~\ref{thm:explicit-final} provides an explicit robustness bound with
directly interpretable dependencies. The prediction deviation scales linearly
with the predictor sensitivity \(L_F\) and only as
\(\sqrt{|\mathcal{C}_{\mathrm{test}}|}\) in the number of test constraints,
matching the natural scaling under uniformly bounded per-constraint embedding
perturbations.

Thus, replacing a test constraint by its nearest neighbor in $\mathcal{F}_{\mathrm{train}}$ induces a provably small change in the constraint-embedding input; by Lipschitz continuity, the corresponding prediction shift is bounded by the stated closed-form expression. More broadly, this yields a direct generalization guarantee across constraint distributions: for constraint families generated by admissible rewrites, the model transfers reliably because the mapping from constraints to embeddings is explicitly stabilized. In other words, the method improves both out-of-distribution constraint generalization and rewrite-robustness by shrinking the only bound-dependent quantity beyond $L_F$ and $|\mathcal{C}_{\mathrm{test}}|$, thereby limiting how much unseen but semantically consistent constraints can change the model input and prediction.

\section{Proofs and Mathematical Background}\label{sec:proof}
\subsection{Technical Lemmas for Section~\ref{sec:4.1}}
\label{sec:lemma_4.1}
\begin{lemma}[Linear constraint embedding]
\label{lem:rigid-linear-zc-form}
Assume Assumption~\ref{assump:linear-encoder-boundedW} and the linear update in Lemma~\ref{lem:bounded-rigid}. Let sum pooling define the constraint embedding $z_c:=\sum_{v\in V_c}h_{v,c}^{(L)}$. Define $S_c:=I_d\otimes\mathbf{1}_{n_c}^\top\in\mathbb{R}^{d\times n_c d}$, where $\otimes$ denotes the Kronecker product and $\mathrm{vec}(\cdot)$ denotes the column-stacking operator. Then
\begin{equation}
z_c=S_c\left(\prod_{\ell=0}^{L-1}T_c^{(\ell)}\right)\mathrm{vec}(H_c^{(0)}),
\end{equation}
where
\begin{equation}
T_c^{(\ell)}=\left(A^{(\ell)}\otimes I_{n_c}\right)+\sum_{r=1}^R\left((B^{(\ell)}W_r^{(\ell)})\otimes A_{c,r}^\top\right).
\end{equation}
\end{lemma}

\begin{lemma}[Stability bound]
\label{lem:rigid-linear-zc-bound}
Under Assumption~\ref{assump:linear-encoder-boundedW} and based on Lemma~\ref{lem:rigid-linear-zc-form}, for any two constraints $c$ and $c'$ that share the same node set size $n:=n_c=n_{c'}$ and the same initialization matrix $H^{(0)}:=H_c^{(0)}=H_{c'}^{(0)}$, their embeddings satisfy
\begin{equation}
\begin{aligned}
&\|z_c-z_{c'}\|_2\le \\
&\|S\|_2\left(\prod_{\ell=0}^{L-1}\Gamma_\ell\right)\|\mathrm{vec}(H^{(0)})\|_2\sum_{\ell=0}^{L-1}\frac{\|T_c^{(\ell)}-T_{c'}^{(\ell)}\|_2}{\Gamma_\ell},
\end{aligned}
\end{equation}
where $S:=I_d\otimes\mathbf{1}_{n}^\top$ and $\Gamma_\ell:=\|T_c^{(\ell)}\|_2\vee \|T_{c'}^{(\ell)}\|_2$. Here $\mathrm{vec}(\cdot)$ denotes the column-stacking operator, and $\vee$ denotes the maximum operator. In particular,
\begin{equation}
\begin{aligned}
\|T_c^{(\ell)}\|_2&\le a_\ell+\sum_{r=1}^R b_\ell w_{\ell,r}\Delta_{c,r},\\
\|T_c^{(\ell)}-T_{c'}^{(\ell)}\|_2&\le \sum_{r=1}^R b_\ell w_{\ell,r}\|A_{c,r}-A_{c',r}\|_2.
\end{aligned}
\end{equation}
\end{lemma}

\subsection{Proof of Lemma~\ref{lem:bounded-rigid}}\label{sec:proof1}
\begin{proof}
By Assumption~\ref{assump:linear-encoder-boundedW}, $\|W_{\mathrm{init}}\|_2\le w_{\mathrm{init}}$ and $\|W_r^{(\ell)}\|_2\le w_{\ell,r}$ for all $\ell,r$.

Substitute the linear instantiations into~\eqref{eq:hl}:
\begin{equation}
h_v^{(\ell+1)}=\psi^{(\ell)} \left(h_v^{(\ell)},\sum_{u:(u,v)\in E_c}\phi^{(\ell)}(h_u^{(\ell)},e_{uv})\right)=A^{(\ell)}h_v^{(\ell)}+B^{(\ell)}\sum_{u:(u,v)\in E_c}W_{r_{uv}}^{(\ell)}h_u^{(\ell)}.
\end{equation}
With $m_{u\to v}^{(\ell)}=W_{r_{uv}}^{(\ell)}h_u^{(\ell)}$, this is exactly
\begin{equation}\label{eq:32}
h_v^{(\ell+1)}=A^{(\ell)}h_v^{(\ell)}+B^{(\ell)}\sum_{u:(u,v)\in E_c}m_{u\to v}^{(\ell)}.
\end{equation}

Define the constants
\begin{equation}
a_\ell:=\|A^{(\ell)}\|_2,\qquad b_\ell:=\|B^{(\ell)}\|_2,\qquad \Delta_{c,r}:=\|A_{c,r}\|_2.
\end{equation}
Then $\|A^{(\ell)}\|_2\le a_\ell$, $\|B^{(\ell)}\|_2\le b_\ell$, and $\|A_{c,r}\|_2\le \Delta_{c,r}$ hold by definition.
\end{proof}

\subsection{Proof of Lemmas~\ref{lem:rigid-linear-zc-form} and~\ref{lem:rigid-linear-zc-bound}}\label{sec:proof2}

\begin{proof}
By Assumption~\ref{assump:linear-encoder-boundedW}, stacking node states row-wise gives $H_c^{(\ell)}\in\mathbb{R}^{n_c\times d}$ with row $v$ equal to $(h_{v,c}^{(\ell)})^\top$. Define $A_{c,r}$ as in Lemma~\ref{lem:bounded-rigid}. For a fixed layer $\ell$, the aggregated message at node $v$ equals
\begin{equation}
    \sum_{u:(u,v)\in E_c}W_{r_{uv}}^{(\ell)}h_{u,c}^{(\ell)}=\sum_{r=1}^R\sum_{u:(u,v)\in E_c, r_{uv}=r}W_r^{(\ell)}h_{u,c}^{(\ell)}.
\end{equation}
Therefore the stacked aggregated message matrix $M_c^{(\ell)}\in\mathbb{R}^{n_c\times d}$ satisfies
\begin{equation}\label{eq:34}
    M_{c,v}^{(\ell)}=\sum_{r=1}^R\left(\sum_{u:(u,v)\in E_c, r_{uv}=r}(h_{u,c}^{(\ell)})^\top\right)(W_r^{(\ell)})^\top
\end{equation}
hence
\begin{equation}\label{eq:35}
M_c^{(\ell)}=\sum_{r=1}^R A_{c,r}^\top H_c^{(\ell)}(W_r^{(\ell)})^\top.
\end{equation}
The stacked node update is $H_c^{(\ell+1)}=H_c^{(\ell)}(A^{(\ell)})^\top+M_c^{(\ell)}(B^{(\ell)})^\top$ by Equation~\eqref{eq:32}, so substituting Equation~\eqref{eq:35} yields
\begin{equation}
H_c^{(\ell+1)}=H_c^{(\ell)}(A^{(\ell)})^\top+\left(\sum_{r=1}^R A_{c,r}^\top H_c^{(\ell)}(W_r^{(\ell)})^\top\right)(B^{(\ell)})^\top.
\end{equation}
Let $h_c^{(\ell)}:=\mathrm{vec}(H_c^{(\ell)})$. By~\cite{horn1991topics} and the identity $\mathrm{vec}(PXQ)=(Q^\top\otimes P)\mathrm{vec}(X)$~\citep{vanloan2000ubiquitous}, we have
\begin{equation}
\begin{aligned}
    \mathrm{vec}(H_c^{(\ell)}(A^{(\ell)})^\top)&=(A^{(\ell)}\otimes I_{n_c})h_c^{(\ell)},\\
    \mathrm{vec}(A_{c,r}^\top H_c^{(\ell)}(W_r^{(\ell)})^\top(B^{(\ell)})^\top)&=((B^{(\ell)}W_r^{(\ell)})\otimes A_{c,r}^\top)h_c^{(\ell)}.
    \end{aligned}
\end{equation}
Summing over $r$ gives
\begin{equation}
h_c^{(\ell+1)}=T_c^{(\ell)}h_c^{(\ell)},\qquad T_c^{(\ell)}=\left(A^{(\ell)}\otimes I_{n_c}\right)+\sum_{r=1}^R\left((B^{(\ell)}W_r^{(\ell)})\otimes A_{c,r}^\top\right).
\end{equation}
Iterating from $\ell=0$ to $L-1$ yields
\begin{equation}
    h_c^{(L)}=\left(\prod_{\ell=0}^{L-1}T_c^{(\ell)}\right)h_c^{(0)}.
\end{equation}
By sum pooling, $z_c=\sum_{v=1}^{n_c}h_{v,c}^{(L)}$, and with $S_c:=I_d\otimes\mathbf{1}_{n_c}^\top$, we have
\begin{equation}
\begin{aligned}
    z_c&=S_c h_c^{(L)}\\
    &=S_c\left(\prod_{\ell=0}^{L-1}T_c^{(\ell)}\right)\mathrm{vec}(H_c^{(0)}).
    \end{aligned}
\end{equation}

We next prove the stability bound between two constraints $c$ and $c'$ with the same $n$ and same $H^{(0)}$. Write $P_c:=\prod_{\ell=0}^{L-1}T_c^{(\ell)}$ and $P_{c'}:=\prod_{\ell=0}^{L-1}T_{c'}^{(\ell)}$. Then
\begin{equation}
    z_c-z_{c'}=S(P_c-P_{c'})\mathrm{vec}(H^{(0)})
\end{equation}
and thus
\begin{equation}\label{eq:42}
\|z_c-z_{c'}\|_2\le \|S\|_2\|P_c-P_{c'}\|_2\|\mathrm{vec}(H^{(0)})\|_2.
\end{equation}
We bound $\|P_c-P_{c'}\|_2$ by a telescoping product inequality: for any sequences $\{X_\ell\}_{\ell=0}^{L-1}$ and $\{Y_\ell\}_{\ell=0}^{L-1}$ of conformable matrices,
\begin{equation}\label{eq:43}
\prod_{\ell=0}^{L-1}X_\ell-\prod_{\ell=0}^{L-1}Y_\ell=\sum_{t=0}^{L-1}\left(\prod_{\ell=t+1}^{L-1}X_\ell\right)(X_t-Y_t)\left(\prod_{\ell=0}^{t-1}Y_\ell\right).
\end{equation}
Applying Equation~\eqref{eq:43} with $X_\ell=T_c^{(\ell)}$ and $Y_\ell=T_{c'}^{(\ell)}$, taking spectral norms, and using the norm bound~\citep{horn1991topics} gives
\begin{equation}
\|P_c-P_{c'}\|_2\le \sum_{t=0}^{L-1}\left(\prod_{\ell=t+1}^{L-1}\|T_c^{(\ell)}\|_2\right)\|T_c^{(t)}-T_{c'}^{(t)}\|_2\left(\prod_{\ell=0}^{t-1}\|T_{c'}^{(\ell)}\|_2\right).
\end{equation}
Let $\Gamma_\ell:=\|T_c^{(\ell)}\|_2\vee \|T_{c'}^{(\ell)}\|_2$. Then each product term is bounded by
\begin{equation}
    \prod_{\ell\ne t}\Gamma_\ell=(\prod_{\ell=0}^{L-1}\Gamma_\ell)/\Gamma_t,
\end{equation}
hence
\begin{equation}
\|P_c-P_{c'}\|_2\le \left(\prod_{\ell=0}^{L-1}\Gamma_\ell\right)\sum_{t=0}^{L-1}\frac{\|T_c^{(t)}-T_{c'}^{(t)}\|_2}{\Gamma_t}.
\end{equation}
Substituting into the bound on $\|z_c-z_{c'}\|_2$ (Equation~\eqref{eq:42}) yields the stated stability inequality.

Finally, we instantiate the norm bounds in terms of parameters and adjacency. By Assumption~\ref{assump:linear-encoder-boundedW} and Lemma~\ref{lem:rigid-linear-zc-form},
\begin{equation}
\begin{aligned}
\|T_c^{(\ell)}\|_2&\le \|A^{(\ell)}\otimes I_n\|_2+\sum_{r=1}^R\|(B^{(\ell)}W_r^{(\ell)})\otimes A_{c,r}^\top\|_2\\
&\le \|A^{(\ell)}\|_2+\sum_{r=1}^R\|B^{(\ell)}\|_2\|W_r^{(\ell)}\|_2\|A_{c,r}\|_2\\
&\le a_\ell+\sum_{r=1}^R b_\ell w_{\ell,r}\Delta_{c,r}.
\end{aligned}
\end{equation}
Moreover, since $A^{(\ell)}$ is shared across $c$ and $c'$,
\begin{equation}
T_c^{(\ell)}-T_{c'}^{(\ell)}=\sum_{r=1}^R\left((B^{(\ell)}W_r^{(\ell)})\otimes (A_{c,r}^\top-A_{c',r}^\top)\right),
\end{equation}
so by the triangle inequality and Kronecker norm multiplicativity~\citep{vanloan2000ubiquitous},
\begin{equation}
\begin{aligned}
\|T_c^{(\ell)}-T_{c'}^{(\ell)}\|_2&\le \sum_{r=1}^R\|B^{(\ell)}W_r^{(\ell)}\|_2\|A_{c,r}-A_{c',r}\|_2\\
&\le \sum_{r=1}^R b_\ell w_{\ell,r}\|A_{c,r}-A_{c',r}\|_2.
\end{aligned}
\end{equation}
This completes the proof.
\end{proof}

\subsection{Proof of Theorem~\ref{thm:one-line-stability}}\label{sec:proof3}

\begin{proof}
Start from Lemma~\ref{lem:rigid-linear-zc-bound}:
\begin{equation}\label{eq:50}
\|z_c-z_{c'}\|_2\le \|S\|_2\left(\prod_{\ell=0}^{L-1}\Gamma_\ell\right)\|\mathrm{vec}(H^{(0)})\|_2\sum_{\ell=0}^{L-1}\frac{\|T_c^{(\ell)}-T_{c'}^{(\ell)}\|_2}{\Gamma_\ell}.
\end{equation}
By Lemma~\ref{lem:rigid-linear-zc-bound} and Assumption~\ref{assump:linear-encoder-boundedW}, with $\Delta_{c,r}=\|A_{c,r}\|_2\le\Delta$ as in Appendix~\ref{sec:proof1}, for every $\ell$,
\begin{equation}
\Gamma_\ell=\|T_c^{(\ell)}\|_2\vee \|T_{c'}^{(\ell)}\|_2\le a+\sum_{r=1}^R bw\Delta=a+bwR\Delta.
\end{equation}
Therefore,
\begin{equation}\label{eq:52}
\prod_{\ell=0}^{L-1}\Gamma_\ell\le (a+bwR\Delta)^L.
\end{equation}
Next, again using Lemma~\ref{lem:rigid-linear-zc-bound} and Assumption~\ref{assump:linear-encoder-boundedW}, for every $\ell$,
\begin{equation}\label{eq:53}
\|T_c^{(\ell)}-T_{c'}^{(\ell)}\|_2\le \sum_{r=1}^R bw\|A_{c,r}-A_{c',r}\|_2\le bwR\delta.
\end{equation}
Hence each summand in Equation~\eqref{eq:50} satisfies
\begin{equation}
\frac{\|T_c^{(\ell)}-T_{c'}^{(\ell)}\|_2}{\Gamma_\ell}\le \frac{bwR\delta}{a+bwR\Delta},
\end{equation}
and summing over $\ell=0,\dots,L-1$ gives
\begin{equation}\label{eq:55}
\sum_{\ell=0}^{L-1}\frac{\|T_c^{(\ell)}-T_{c'}^{(\ell)}\|_2}{\Gamma_\ell}\le \frac{L\cdot bwR\delta}{a+bwR\Delta}.
\end{equation}
Plugging~\eqref{eq:52} and~\eqref{eq:55} into the stability inequality~\eqref{eq:50} yields
\begin{equation}
\|z_c-z_{c'}\|_2\le \|S\|_2\|\mathrm{vec}(H^{(0)})\|_2(a+bwR\Delta)^{L}\cdot \frac{L\cdot bwR\delta}{a+bwR\Delta}.
\end{equation}
Cancelling one factor of $(a+bwR\Delta)$ gives the equivalent one-line form
\begin{equation}
\|z_c-z_{c'}\|_2\le \|S\|_2\|\mathrm{vec}(H^{(0)})\|_2\cdot L\cdot bwR\delta\cdot (a+bwR\Delta)^{L-1}.
\end{equation}
\end{proof}

\subsection{Discussion of Assumption~\ref{assump:linear-encoder-boundedW}}\label{app:assumption-discussion}
Assumption~\ref{assump:linear-encoder-boundedW} follows the linearized message-passing form standard in MPNN stability analysis~\citep{gilmer2017mpnn, zhou2020gnn}, with bounded parameters required for any meaningful perturbation bound~\citep{stewart1990matrix, higham2002accuracy}. The trained encoder uses ReLU, LayerNorm, and attention pooling beyond this idealization; empirical stability under reparameterization is shown by Table~\ref{tab:sensitivity}. The same argument extends to this nonlinear encoder when the layer maps are Lipschitz: replacing each per-layer spectral norm by the corresponding Lipschitz constant gives a bound of the same form. Table~\ref{tab:sensitivity} measures the stability of the trained encoder directly.

\subsection{Proof of Lemma~\ref{lem:loss-to-sim}}\label{sec:proof4}

\begin{proof}
Write $z=z_c$, $z^+=z_{c^+}$, and $z_k^-=z_{c_k^-}$. Define the scaled similarities $s^+=\mathrm{sim}(z,z^+)/\tau$ and $s_k^-=\mathrm{sim}(z,z_k^-)/\tau$. By Equation~\eqref{eq:loss}, the contrastive loss can be rewritten as
\begin{equation}
\mathcal{L}_{\mathrm{con}}(c)=-\log\left(\frac{e^{s^+}}{e^{s^+}+\sum_{k=1}^K e^{s_k^-}}\right).
\end{equation}
Define
\begin{equation}\label{eq:59}
    p_{\mathrm{pos}}=\frac{e^{s^+}}{e^{s^+}+\sum_{k=1}^K e^{s_k^-}}.
\end{equation}
Then $\mathcal{L}_{\mathrm{con}}(c)=-\log p_{\mathrm{pos}}$. The condition $\mathcal{L}_{\mathrm{con}}(c)\le \varepsilon$ implies
\begin{equation}\label{eq:60}
p_{\mathrm{pos}}\ge e^{-\varepsilon}.
\end{equation}
Invert $p_{\mathrm{pos}}$ to isolate the negative partition term in~\eqref{eq:59}:
\begin{equation}
\frac{1}{p_{\mathrm{pos}}}=1+\sum_{k=1}^K e^{s_k^- - s^+}.
\end{equation}
Combining with~\eqref{eq:60} yields
\begin{equation}
1+\sum_{k=1}^K e^{s_k^- - s^+}\le e^\varepsilon,
\end{equation}
and hence
\begin{equation}
\sum_{k=1}^K e^{s_k^- - s^+}\le e^\varepsilon-1.
\end{equation}
By the assumption in Lemma~\ref{lem:loss-to-sim}, $\mathrm{sim}(z,z_k^-)\ge \rho^-$ for every $k=1,\dots,K$, so $s_k^-\ge \rho^-/\tau$. Therefore each summand satisfies $e^{s_k^- - s^+}\ge e^{\rho^-/\tau - s^+}$, and summing over $k$ yields
\begin{equation}\label{eq:64}
\sum_{k=1}^K e^{s_k^- - s^+}\ge K e^{\rho^-/\tau - s^+}.
\end{equation}
Combining~\eqref{eq:64} with $\sum_{k=1}^K e^{s_k^- - s^+}\le e^\varepsilon-1$ gives
\begin{equation}
K e^{\rho^-/\tau - s^+}\le e^\varepsilon-1.
\end{equation}
Taking logarithms and rearranging yields
\begin{equation}
s^+\ge \frac{\rho^-}{\tau}+\log K-\log(e^\varepsilon-1).
\end{equation}
Finally, substituting $s^+=\mathrm{sim}(z,z^+)/\tau$ gives
\begin{equation}
\mathrm{sim}(z_c,z_{c^+})\ge \rho^-+\tau\left(\log K-\log(e^\varepsilon-1)\right).
\end{equation}
\end{proof}

\subsection{Proof of Lemma~\ref{lem:coverage}}\label{sec:proof5}
To rigorously prove Lemma~\ref{lem:coverage}, we first prove Lemma~\ref{lem:sim-to-dist} and Lemma~\ref{lem:loss-to-dist}.

\begin{lemma}[From cosine similarity to an $\ell_2$ distance identity]
\label{lem:sim-to-dist}
Under the assumptions of Lemma~\ref{lem:coverage}, for any pair $(c,c')$, $\|z_c-z_{c'}\|_2^2=2\left(1-\mathrm{sim}(z_c,z_{c'})\right)$.
\end{lemma}

\begin{proof}
By the $\ell_2$-normalization assumption in Lemma~\ref{lem:coverage}, $\|z_c\|_2=\|z_{c'}\|_2=1$. Expand the squared distance:
\begin{equation}
\begin{aligned}
\|z_c-z_{c'}\|_2^2 &=\|z_c\|_2^2+\|z_{c'}\|_2^2-2\langle z_c,z_{c'}\rangle\\
&=2-2\langle z_c,z_{c'}\rangle.
\end{aligned}
\end{equation}
Since cosine similarity equals the inner product for $\ell_2$-normalized embeddings, $\langle z_c,z_{c'}\rangle=\mathrm{sim}(z_c,z_{c'})$. Substitute to obtain the claim.
\end{proof}

\begin{lemma}[From contrastive loss to an explicit bound on reparameterization embedding distance]
\label{lem:loss-to-dist}
Fix a constraint $c$ and its positive $c^+=T(c)$ with $T\in\mathcal{T}$. Under the assumptions of Lemma~\ref{lem:coverage}, if $\mathcal{L}_{\mathrm{con}}(c)\le \varepsilon$, then
\begin{equation}
\|z_c-z_{c^+}\|_2^2\le 2\left(1-\rho^--\tau\left(\log K-\log(e^\varepsilon-1)\right)\right).
\end{equation}
\end{lemma}

\begin{proof}
By Lemma~\ref{lem:loss-to-sim}, $\mathrm{sim}(z_c,z_{c^+})\ge \rho^-+\tau\left(\log K-\log(e^\varepsilon-1)\right)$. Apply Lemma~\ref{lem:sim-to-dist} with $c'=c^+$:
\begin{equation}
\|z_c-z_{c^+}\|_2^2=2\left(1-\mathrm{sim}(z_c,z_{c^+})\right).
\end{equation}
Substitute the lower bound on $\mathrm{sim}(z_c,z_{c^+})$ into the right-hand side:
\begin{equation}
\|z_c-z_{c^+}\|_2^2\le 2\left(1-\rho^--\tau\left(\log K-\log(e^\varepsilon-1)\right)\right).
\end{equation}
\end{proof}

We now prove Lemma~\ref{lem:coverage}.

\begin{proof}
Fix any $c\in\mathcal{F}$. By Assumption~\ref{assump:orbit-cover}, there exist $c'\in\mathcal{F}_{\mathrm{train}}$ and $T\in\mathcal{T}$ such that $c=T(c')$. By the definition of $\Delta_\phi$,
\begin{equation}\label{eq:78}
\begin{aligned}
\Delta_\phi(c;\mathcal{F}_{\mathrm{train}})&=\min_{\bar c\in\mathcal{F}_{\mathrm{train}}}\|\phi_\theta(c)-\phi_\theta(\bar c)\|_2\\
&\le \|\phi_\theta(c)-\phi_\theta(c')\|_2.
\end{aligned}
\end{equation}
Since $c=T(c')$, the right-hand side equals
\begin{equation}
\|\phi_\theta(T(c'))-\phi_\theta(c')\|_2=\|z_{T(c')}-z_{c'}\|_2.
\end{equation}
Now use the training-time condition $\mathcal{L}_{\mathrm{con}}(c',T(c'))\le \varepsilon$. By construction of positives in the contrastive objective, $T(c')$ is a valid positive for anchor $c'$ (same admissible set $\mathcal{T}$), so Lemma~\ref{lem:loss-to-dist} applied to anchor $c'$ yields
\begin{equation}
\|z_{c'}-z_{T(c')}\|_2^2\le 2\left(1-\rho^--\tau\left(\log K-\log(e^\varepsilon-1)\right)\right).
\end{equation}
Combine with~\eqref{eq:78} to obtain
\begin{equation}
\Delta_\phi(c;\mathcal{F}_{\mathrm{train}})\le \sqrt{2\left(1-\rho^--\tau\left(\log K-\log(e^\varepsilon-1)\right)\right)}.
\end{equation}
Because this holds for every $c\in\mathcal{F}$, take $\sup_{c\in\mathcal{F}}$ on the left-hand side, which by Definition~\ref{def:coverage} equals $R_\phi(\mathcal{F};\mathcal{F}_{\mathrm{train}})$, yielding the stated bound.
\end{proof}

\subsection{Proof of Lemma~\ref{lem:gen-by-coverage}}\label{sec:proof6}

\begin{proof}
Let $m=|\mathcal{C}_{\mathrm{test}}|$. Fix an ordering $\mathcal{C}_{\mathrm{test}}=\{c_1,\dots,c_m\}$. For each $i$, let $c_i^{\mathrm{nn}}\in\mathcal{F}_{\mathrm{train}}$ be a nearest neighbor in embedding space, meaning
\begin{equation}
\|z_{c_i}-z_{c_i^{\mathrm{nn}}}\|_2=\Delta_\phi(c_i;\mathcal{F}_{\mathrm{train}}).
\end{equation}
Stack embeddings row-wise as $Z_{\mathcal{C}_{\mathrm{test}}}\in\mathbb{R}^{m\times d}$ with $i$-th row $z_{c_i}^\top$ and $Z_{\mathcal{C}_{\mathrm{nn}}}\in\mathbb{R}^{m\times d}$ with $i$-th row $z_{c_i^{\mathrm{nn}}}^\top$. By Assumption~\ref{assump:lipschitz},
\begin{equation}
\begin{aligned}
\|\hat x_{\mathrm{test}}-\hat x_{\mathrm{nn}}\|&=\|F_\theta(\mathcal{G},f,Z_{\mathcal{C}_{\mathrm{test}}})-F_\theta(\mathcal{G},f,Z_{\mathcal{C}_{\mathrm{nn}}})\|\\
&\le L_F\|Z_{\mathcal{C}_{\mathrm{test}}}-Z_{\mathcal{C}_{\mathrm{nn}}}\|_F.
\end{aligned}
\end{equation}
It remains to bound $\|Z_{\mathcal{C}_{\mathrm{test}}}-Z_{\mathcal{C}_{\mathrm{nn}}}\|_F$. By definition of the Frobenius norm,
\begin{equation}
\|Z_{\mathcal{C}_{\mathrm{test}}}-Z_{\mathcal{C}_{\mathrm{nn}}}\|_F^2=\sum_{i=1}^m \|z_{c_i}-z_{c_i^{\mathrm{nn}}}\|_2^2.
\end{equation}
By Definition~\ref{def:coverage}, for each $i$ we have
\begin{equation}
\begin{aligned}
\|z_{c_i}-z_{c_i^{\mathrm{nn}}}\|_2&=\Delta_\phi(c_i;\mathcal{F}_{\mathrm{train}})\\
&\le R_\phi(\mathcal{F};\mathcal{F}_{\mathrm{train}}).
\end{aligned}
\end{equation}
Therefore each term in the sum is at most $R_\phi(\mathcal{F};\mathcal{F}_{\mathrm{train}})^2$, giving
\begin{equation}
\begin{aligned}
\|Z_{\mathcal{C}_{\mathrm{test}}}-Z_{\mathcal{C}_{\mathrm{nn}}}\|_F^2&\le \sum_{i=1}^m R_\phi(\mathcal{F};\mathcal{F}_{\mathrm{train}})^2\\
&=m R_\phi(\mathcal{F};\mathcal{F}_{\mathrm{train}})^2.
\end{aligned}
\end{equation}
Taking square roots yields
\begin{equation}\label{eq:77}
\|Z_{\mathcal{C}_{\mathrm{test}}}-Z_{\mathcal{C}_{\mathrm{nn}}}\|_F\le \sqrt{|\mathcal{C}_{\mathrm{test}}|} R_\phi(\mathcal{F};\mathcal{F}_{\mathrm{train}}).
\end{equation}
Substitute~\eqref{eq:77} into Assumption~\ref{assump:lipschitz} to obtain the claimed result.
\end{proof}

\subsection{Proof of Theorem~\ref{thm:explicit-final}}\label{sec:proof7}

\begin{proof}
By Lemma~\ref{lem:gen-by-coverage},
\begin{equation}
\|F_\theta(\mathcal{G},f,Z_{\mathcal{C}_{\mathrm{test}}})-F_\theta(\mathcal{G},f,Z_{\mathcal{C}_{\mathrm{nn}}})\|\le L_F\sqrt{|\mathcal{C}_{\mathrm{test}}|}\,R_\phi(\mathcal{F};\mathcal{F}_{\mathrm{train}}).
\end{equation}
Applying the coverage-radius bound of Lemma~\ref{lem:coverage} yields
\begin{equation}
\begin{aligned}
&\|F_\theta(\mathcal{G},f,Z_{\mathcal{C}_{\mathrm{test}}})-F_\theta(\mathcal{G},f,Z_{\mathcal{C}_{\mathrm{nn}}})\| \\
&\le L_F\sqrt{|\mathcal{C}_{\mathrm{test}}|}\sqrt{2\left(1-\rho^--\tau\left(\log K-\log(e^\varepsilon-1)\right)\right)}.
\end{aligned}
\end{equation}
\end{proof}

\section{Datasets and Implementation Details}
\subsection{Datasets}\label{subsec:datasets}
\paragraph{MILP datasets.}
For MILP problems, we use datasets from the GOAL paper. Each task contains 1{,}000{,}000 training instances, for a total of 8 million instances across tasks. Standard tasks use graphs with 100 nodes, while JSSP and UMSP use 10$\times$10 and 100$\times$20 operation layouts (100 operation nodes), respectively. Supervision is provided via full imitation learning from expert trajectories generated by classical solvers. A validation set of 128 instances is used for each task.

\paragraph{Grid optimization datasets.}
For power-grid optimization tasks, datasets are generated across different formulations of the underlying network constraints, reflecting multiple ways of representing the same physical problems within \textsc{PowerModels}. Specifically, we generate 15{,}000 samples per case for ACPF, DCPF, and ACOPF formulations, and 5{,}000 samples per case for DCOPF, DCOTS, and DCTNEP. All instances are constructed with \textsc{PowerModels} and solved with traditional optimization solvers to obtain supervision.

Tables~\ref{tab:power_grid_overview} and~\ref{tab:problem_overview} summarize the power-grid and combinatorial optimization problems used in our experiments, including their constraint structures, representative formulations, and problem classes.

Tables~\ref{tab:milp_node_edge_features} and~\ref{tab:opf_node_edge_features} detail the corresponding node and edge feature representations used for graph-based modeling in combinatorial and power-grid domains, respectively. The feature specifications are included for completeness and to clarify the input representations used across different problem families.

\begin{table}[t]
\centering
\renewcommand{\arraystretch}{1.25}
\caption{Power-grid optimization problems with their constraint structures, representative equations, and mathematical classes.}
\label{tab:power_grid_overview}
\resizebox{\textwidth}{!}{
\begin{tabular}{p{3.4cm} p{4.6cm} p{5.2cm} p{2.0cm} p{5.2cm}}
\hline
\textbf{Problem}
& \textbf{Constraint Types}
& \textbf{Representative Equations}
& \textbf{Class}
& \textbf{Description} \\
\hline

AC Power Flow (AC PF)
& AC power balance; voltage magnitude limits; branch Ohm's law
& $P_i = \sum_{j} V_i V_j (G_{ij}\cos\theta_{ij} + B_{ij}\sin\theta_{ij})$;
$V_i^{\min} \le |V_i| \le V_i^{\max}$;
$I_{ij} = Y_{ij}(V_i - V_j)$
& NLP
& Solve nonlinear AC network physics (voltages, injections, flows). \\ \hline

DC Power Flow (DC PF)
& Linearized power flow; nodal power balance
& $P_{ij} = B_{ij}(\theta_i - \theta_j)$;
$P_i = \sum_j P_{ij}$
& LP
& Linear active-power-only approximation of power flow. \\ \hline

AC Optimal Power Flow (AC OPF)
& AC power balance; generator limits; voltage limits; thermal limits
& $P_i = \sum_{j} V_i V_j (G_{ij}\cos\theta_{ij} + B_{ij}\sin\theta_{ij})$;
$P_g^{\min} \le P_g \le P_g^{\max}$;
$V_i^{\min} \le |V_i| \le V_i^{\max}$;
$|S_{ij}| \le S_{ij}^{\max}$
& NLP
& Minimize generation cost subject to full nonlinear AC network constraints. \\ \hline

DC Optimal Power Flow (DC OPF)
& Linear DC power flow; generator limits; line thermal limits
& $P_{ij} = B_{ij}(\theta_i - \theta_j)$;
$P_g^{\min} \le P_g \le P_g^{\max}$;
$|P_{ij}| \le P_{ij}^{\max}$
& LP/QP
& Economic dispatch with linearized DC network and flow limits. \\ \hline

DC Optimal Transmission Switching (DC OTS)
& Linear DC power flow; line-status binaries; thermal limits; generator limits
& $P_{ij} = B_{ij}(\theta_i - \theta_j)x_{ij}$;
$x_{ij}\in\{0,1\}$;
$|P_{ij}| \le x_{ij} P_{ij}^{\max}$;
$P_g^{\min} \le P_g \le P_g^{\max}$
& MILP
& DC OPF with binary switching of transmission elements. \\ \hline

DC Transmission Network Expansion Planning (DC TNEP)
& DC power flow; build binaries; thermal limits; budget constraint; generator limits
& $P_{ij} = B_{ij}(\theta_i - \theta_j)y_{ij}$;
$y_{ij}\in\{0,1\}$;
$|P_{ij}| \le y_{ij} P_{ij}^{\max}$;
$\sum_{(i,j)} c_{ij} y_{ij} \le \mathrm{Budget}$
& MILP
& Decide which DC lines to build or upgrade under budget and DC network constraints. \\ \hline

\end{tabular}
}
\end{table}

\begin{table}[t]
\centering
\renewcommand{\arraystretch}{1.25}
\caption{Overview of combinatorial optimization problems, their constraint structures, objectives, and formulations.}
\label{tab:problem_overview}
\resizebox{\textwidth}{!}{
\begin{tabular}{p{3.2cm} p{4.2cm} p{3.6cm} p{2.0cm} p{5.5cm}}
\hline
\textbf{Problem}
& \textbf{Constraint Types}
& \textbf{Objective Function}
& \textbf{Type}
& \textbf{Description} \\
\hline

ATSP
& Degree; Subtour elimination
& $\displaystyle \min \sum_{(i,j)} c_{ij} x_{ij}$
& ILP/MILP
& Find minimum-cost Hamiltonian tour in directed graph. \\ \hline

CVRP
& Flow; Capacity; Subtour elimination
& $\displaystyle \min \sum_{(i,j)} c_{ij} x_{ij}$
& MILP
& Deliver to all customers with limited vehicle capacity. \\ \hline

CVRPTW
& Flow; Capacity; Time windows; Time propagation
& Same as CVRP
& MILP
& Serve all customers respecting time-window constraints. \\ \hline

OP
& Flow; Budget; Reward linking; Subtour elimination
& $\displaystyle \max \sum_i p_i y_i$
& MILP
& Maximize collected reward under travel budget. \\ \hline

Knapsack (KP)
& Weight constraint
& $\displaystyle \max \sum_i v_i x_i$
& ILP
& Select items maximizing value under total weight. \\ \hline

MVC
& Edge covering: $x_u + x_v \ge 1$
& $\displaystyle \min \sum_i x_i$
& ILP
& Choose minimum vertices covering all edges. \\ \hline

UMSP
& Assignment; Load constraints
& $\displaystyle \min C_{\max}$
& MILP
& Assign jobs to unrelated machines with different processing times. \\ \hline

JSSP
& Precedence; No-overlap; Time propagation
& $\displaystyle \min C_{\max}$
& MILP
& Sequence job operations on machines with strict order. \\ \hline

TRP
& Flow; Arrival propagation; Subtour elimination (variant)
& $\displaystyle \min \sum_i \mathrm{arrival}_i$
& MILP
& Minimize total waiting time of all visited nodes. \\ \hline

PCTSP
& Flow; Reward linking; Subtour elimination
& $\displaystyle \min \sum c_{ij}x_{ij} - \sum_i p_i y_i$
& MILP
& Visit profitable nodes while minimizing travel cost. \\ \hline

OCVRP
& Flow; Capacity; No-return constraint
& $\displaystyle \min \sum c_{ij}x_{ij}$
& MILP
& Vehicles end at last customer (no depot return). \\ \hline

SDCVRP
& Flow; Capacity; Split-delivery constraints
& Same as CVRP
& MILP
& Allow customers to receive split deliveries. \\ \hline

SOP
& Precedence; Flow; Subtour elimination
& $\displaystyle \min \sum c_{ij} x_{ij}$
& MILP/ILP
& TSP with explicit node-precedence requirements. \\ \hline

MCLP
& Facility budget; Coverage linking
& $\displaystyle \max \sum_j w_j z_j$
& ILP
& Place facilities to maximize demand coverage. \\ \hline

MIS
& Independence: $x_u + x_v \le 1$
& $\displaystyle \max \sum_i x_i$
& ILP
& Select vertices with no edges between them. \\ \hline

OSSP
& No-overlap; Task assignment
& $\displaystyle \min C_{\max}$
& MILP
& Schedule each job on all machines without ordering. \\ \hline

\end{tabular}
}
\end{table}

\begin{table}[t]
\centering
\renewcommand{\arraystretch}{1.15}
\setlength{\tabcolsep}{6pt}
\caption{Node and edge feature representations across MILP problems.}
\label{tab:milp_node_edge_features}
\resizebox{\textwidth}{!}{
\begin{tabular}{l l p{5.6cm} p{5.6cm}}
\hline
\textbf{Problem} & \textbf{Node Type} & \textbf{Node Features} & \textbf{Edge Features} \\
\hline

ATSP
& City
& Random ID feature (symmetry breaking)
& Asymmetric distance / cost matrix \\
\hline

CVRP
& Customer / Depot
& Demand, depot indicator, random ID
& Distance / travel cost \\
\hline

CVRPTW
& Customer / Depot
& Demand, time window $[e_i,l_i]$, service time, depot flag
& Distance / travel time \\
\hline

OP (Orienteering)
& Location
& Prize / reward, random ID
& Distance / travel cost \\
\hline

KP (Knapsack)
& Item
& Weight, value
& -- \\
\hline

MVC (Min Vertex Cover)
& Graph node
& Random ID, optional degree
& Binary adjacency \\
\hline

UMSP
& Operation
& Processing time, job ID, random ID
& Machine--operation compatibility, processing cost \\
\hline

JSSP
& Operation / Machine
& Operation duration, job index, machine index
& Precedence constraints, machine assignment \\
\hline

TRP
& City
& Random ID
& Distance / travel cost \\
\hline

PCTSP
& City
& Prize, penalty for skipping
& Distance / travel cost \\
\hline

OCVRP
& Customer
& Demand, random ID
& Distance / travel cost \\
\hline

SDCVRP
& Customer
& Demand, split-allowed flag
& Distance / travel cost \\
\hline

SOP
& City
& Random ID
& Precedence-constrained distance matrix \\
\hline

MCLP
& Location
& Facility indicator, coverage radius
& Coverage relation (binary / distance-based) \\
\hline

MIS (Max Independent Set)
& Graph node
& Random ID, optional degree
& Binary adjacency \\
\hline

OSSP
& Operation / Machine
& Processing time, job ID
& Machine--operation compatibility \\
\hline

\end{tabular}
}
\end{table}

\begin{table}[t]
\centering
\setlength{\tabcolsep}{6pt}
\renewcommand{\arraystretch}{1.15}
\caption{Node and edge features used in power grid optimization problems. AC formulations model full complex power flow physics, while DC variants use linearized active power flow and may include discrete line switching or expansion decisions.}
\label{tab:opf_node_edge_features}
\begin{tabular}{p{2.2cm} p{5.2cm} p{5.2cm}}
\hline
\textbf{Problem} & \textbf{Node Features} & \textbf{Edge Features} \\
\hline
ACPF
& Bus voltage bounds $(v^{\min}, v^{\max})$, bus type (PQ/PV/Slack), base kV;
Load demand $(p_d, q_d)$;
Fixed generator injections $(p_g, q_g)$
& Line/transformer resistance $r$, reactance $x$, line charging $(b_{fr}, b_{to})$, thermal limits $s^{\max}$, angle difference bounds $(\theta^{\min}, \theta^{\max})$, tap ratio, phase shift \\
\hline
ACOPF
& Bus voltage bounds and type;
Generator limits $(p_g^{\min}, p_g^{\max}, q_g^{\min}, q_g^{\max})$, cost coefficients $(c_2, c_1, c_0)$, voltage setpoint $v_g$;
Load $(p_d, q_d)$;
Shunt parameters $(g_s, b_s)$
& Line/transformer parameters $r, x$, charging susceptance, thermal limits (rate A/B/C), tap ratio, phase shift, angle difference bounds \\
\hline
DCOPF
& Reference bus indicator;
Generator limits $(p_g^{\min}, p_g^{\max})$ and cost;
Active load $p_d$
& Line susceptance $b = 1/x$ and thermal limits $p^{\max}$ \\
\hline
DCOTS
& Reference bus indicator;
Generator limits and cost;
Active load $p_d$
& Line susceptance and thermal limits;
binary line switching variable (on/off) \\
\hline
DCTNEP
& Reference bus indicator;
Generator limits and cost;
Active load $p_d$
& Candidate line susceptance and thermal limits;
binary line construction variable with investment cost \\
\hline
\end{tabular}
\end{table}

\subsection{Implementation Details}
\label{subsec:imple_details}
\paragraph{Training.}
For grid optimization problems, the model is trained for 300 epochs.
For combinatorial optimization tasks formulated as MILPs, training is performed for 450 epochs to account for the increased complexity of discrete decision spaces and decoding.
Experiments are conducted on NVIDIA GPUs.

\paragraph{Key architectural details.}
CONSTRAINER employs a Heterogeneous Graph Transformer (HGT) as the shared backbone for both grid and MILP problems, enabling unified processing of heterogeneous node and edge types.
For continuous optimization problems, task-specific decoders are implemented as lightweight MLP heads that predict output variables associated with graph nodes or edges (e.g., generator dispatch or line flows).
For MILP problems, we adopt a greedy decoding procedure identical to that used in GOAL, and apply the same decoding strategy consistently across all baselines and CONSTRAINER.

\paragraph{Baseline adaptation.}
For OPTFM, task-specific heads are finetuned on top of the shared MILP representation to adapt to each target problem.
For the multi-task learning (MTL) baseline, separate input--output heads are used for each task while sharing the backbone encoder.
For single-task grid optimization problems where naive GNN or GAT baselines are not directly applicable, we adapt the ACOPF baseline by introducing edge-level prediction heads for DCOTS and DCTNEP, treating line switching or construction as edge selection tasks.

\paragraph{Training objectives.} For MILP tasks, pretraining uses an imitation-learning objective (per-step cross-entropy on solver-optimal trajectories), following GOAL~\citep{goal2025}, and finetuning uses an unsupervised self-imitation objective that samples candidate solutions and reinforces high-quality trajectories, without oracle labels. For grid tasks, the objective is a supervised regression loss against the solver solution augmented with a constraint-violation penalty. The constraint encoder is trained separately with the contrastive objective in Eq.~\eqref{eq:loss}.

\paragraph{Reference solvers.} Grid labels are generated with IPOPT (PF, ACPF, ACOPF) and HiGHS (DCOPF, DCOTS, DCTNEP). The combinatorial labels are the public GOAL datasets, solved by the per-task oracle solvers of \citet{goal2025}: LKH (ATSP, CVRP), HGS (CVRPTW), A4OP (OP), OR-Tools (KP, JSSP), FastWVC (MVC), and HiGHS (UMSP).

\paragraph{Baseline provenance.} The generalist foundation-model baselines (OPTFM, MTL, GOAL) are evaluated on all tasks under our unified protocol. The specialist baselines are single-task models, each built for one problem type; a dash in the results tables marks a task outside a baseline's scope.

\paragraph{Finetuning details.}
For MILP problems, we perform unsupervised finetuning without using oracle labels, using a self-imitation strategy based on sampling candidate solutions and reinforcing high-quality trajectories, following prior practice in learning-based combinatorial optimization.

For grid optimization problems, finetuning is performed in a supervised manner using ground-truth solutions obtained from conventional solvers, optimizing prediction accuracy while enforcing feasibility through task-specific constraints.

\paragraph{Evaluation metrics.}
We report feasibility and optimality gap. Feasibility is defined as the percentage of test instances for which the predicted solution satisfies all problem constraints within predefined solver tolerances. The optimality gap is computed as
\[
\text{Gap}(\%) = \frac{|f(x_{\text{pred}}) - f(x^\star)|}{|f(x^\star)|} \times 100,
\]
where $f(x_{\text{pred}})$ denotes the objective value of the predicted solution, and $f(x^\star)$ is the objective value of the corresponding optimal solution obtained from a classical solver.

\paragraph{Numerical handling of constants.}
Scalar constant values $x$ are passed through a signed log transform $\tilde{x} = \mathrm{sign}(x) \cdot \log(1 + |x|)$ before the MLP projection $\phi(\cdot)$. This compresses very large values (e.g., $10^{6} \mapsto 13.8$) while leaving small values essentially unchanged (e.g., $10^{-3} \mapsto 10^{-3}$) and preserves sign and ordering, so $\phi(\tilde{x})$ remains well-conditioned across heterogeneous problem domains.

\paragraph{Computational cost.}
We measure training and per-instance inference time on CVRP at 100 nodes with training batch size 64 and inference batch size 1, finetuned for 80 epochs (Table~\ref{tab:timing_cvrp}). CONSTRAINER trains slightly faster than OPTFM and matches GOAL's per-instance inference time, with no statistically significant overhead from the constraint cross-attention module. At inference, constraint embeddings are cached, so the per-instance cost of CONSTRAINER reduces to that of the shared backbone.

\begin{table}[t]
\centering
\small
\setlength{\tabcolsep}{6pt}
\renewcommand{\arraystretch}{1.1}
\caption{Training and per-instance inference time on CVRP, 100 nodes.}
\label{tab:timing_cvrp}
\begin{tabular}{l|c|c}
\hline
Method & Training time (hr) & Inference (ms / instance) \\
\hline
OPTFM       & 0.31 & 1230 \\
GOAL        & 0.32 & 334.6 $\pm$ 23.8 \\
CONSTRAINER & \textbf{0.26} & \textbf{325.0} $\pm$ 17.6 \\
\hline
\end{tabular}
\end{table}

To assess scaling with graph size, we additionally measure inference time of CONSTRAINER on ATSP at 100, 200, and 500 nodes (Table~\ref{tab:timing_atsp_scale}).
The per-instance cost decomposes by module. The constraint encoder costs $O(\sum_{c}|E_c|\,d)$; its output is cached at inference, so its per-instance cost is $O(1)$. The backbone costs $O(L\,|E|\,d)$, and for the dense ATSP graph $|E|=O(n^2)$ in the number of cities $n$. Constraint cross-attention costs $O(|V|\,|\mathcal{C}|\,d)$, linear in $|V|$ for the small constraint sets used here ($|\mathcal{C}|\in[1,7]$). At the sizes in Table~\ref{tab:timing_atsp_scale}, inference time is dominated by fixed per-call overhead; the growth with graph size comes from the backbone, and the conditioning module adds no measurable overhead.

\begin{table}[t]
\centering
\small
\setlength{\tabcolsep}{6pt}
\renewcommand{\arraystretch}{1.1}
\caption{Inference scaling of CONSTRAINER on ATSP across graph sizes.}
\label{tab:timing_atsp_scale}
\begin{tabular}{l|c}
\hline
Problem size (nodes) & Inference time (ms) \\
\hline
100 & 60.1 \\
200 & 69.2 \\
500 & 345.8 \\
\hline
\end{tabular}
\end{table}

\paragraph{Contrastive pair construction details.}
Table~\ref{tab:contras} lists constructions of positive and negative pairs for different constraint categories used in the contrastive learning setup. These illustrate the types of invariances and distinctions considered during training of the constraint encoder.
\begin{table}[t]
\centering
\renewcommand{\arraystretch}{1.25}
\scriptsize
\caption{Geometric characterization of major constraint types and how to construct positive/negative pairs for contrastive learning of a constraint encoder.}
\label{tab:contras}
\resizebox{\textwidth}{!}{
\begin{tabular}{|p{2.3cm}|p{2.6cm}|p{3.0cm}|p{2.8cm}|p{3.6cm}|p{3.6cm}|}
\hline
\textbf{Constraint Type} &
\textbf{Geometric Object} &
\textbf{Invariances to Learn} &
\textbf{Contrastive Signal} &
\textbf{Positive Pair Construction} &
\textbf{Negative Pair Construction} \\ \hline

Linear Equality
& Hyperplane $a^\top x = b$
& Scaling invariance; parallelism; translation; redundancy
& Identify equivalent hyperplanes
& Rescale $(a,b)$; add redundant copies; reorder variables
& Rotate normal; shift $b$; inconsistent systems \\ \hline

Linear Inequality
& Half-space $a^\top x \le b$
& Tightness ordering; half-space orientation; subset relations
& Ordered distances based on looseness/tightness
& Generate looser/tighter $b'$; same normal direction
& Reverse inequality; change normal; produce disjoint regions \\ \hline

Affine Constraint
& Shifted subspace
& Affine transform invariance; equivalence of same geometry
& Close embeddings for affine-equivalent forms
& Apply $x \mapsto Ax+c$; add dummy variables
& Non-parallel normals; incompatible affine offsets \\ \hline

Box Constraint
& Hyperrectangle $l_i \le x_i \le u_i$
& Nesting; axis-aligned invariance; independent dims
& Encode bounding hierarchy
& Nested boxes; reorder dimensions
& Disjoint boxes; rotated boxes; contradictory bounds \\ \hline

Integer Constraint
& Lattice $\mathbb{Z}^n$
& Relaxation equivalence; parity structure; modular patterns
& Integer $\approx$ relaxed hull (close but distinct)
& Integer vs continuous versions; parity vs equivalent parity
& Arbitrary continuous sets; incompatible mod classes \\ \hline

Logical Big-M
& Union of polyhedra; disjunctions
& Branch invariance; M-insensitivity; piecewise geometry
& Capture branch semantics rather than algebra
& Same implication w/ different $M$; reordering binary vars
& Different logical semantics; huge $M$ destroying geometry \\ \hline

Ordering Constraint
& Monotone cone ($x_i \le x_j$)
& Poset invariance; chain equivalence; cone orientation
& Group constraints forming same order structure
& Same poset under renaming; add transitive closures
& Cyclic orders; inconsistent chains; reversed inequalities \\ \hline
\end{tabular}
}
\end{table}

\subsection{Scope of representable constraints}\label{app:scope}
The DAG encoding represents any constraint expressible as a finite computation graph over the framework's operator vocabulary. The vocabulary instantiated in this paper includes arithmetic operators (e.g., $+$, $\times$, $\mathrm{pow}$), comparison operators ($\le$, $=$, $\ge$), and is extensible: additional operators can be added by introducing new primitive types in the encoder lookup table $\phi_{\mathrm{op}}$ and exposing them in the parser. This covers linear, quadratic, polynomial, and algebraic nonlinear constraints used throughout the MILP, QP, and OPF formulations evaluated in the paper. Constraints with compact algebraic reformulations, including flow- and cut-based encodings of structural properties, fall within this class via their reformulations. The framework adapts to new problem domains through a localized update: when new operators arise, only the encoder lookup $\phi_{\mathrm{op}}$ is retrained on examples containing them, while the backbone, cross-attention, and task decoder remain unchanged.

\subsection{Additional Experimental Results}
\paragraph{Additional grid-scale results.}
Table~\ref{tab:grid_generalization} summarizes additional results of CONSTRAINER on OPF problems across larger grid instances, including IEEE benchmark systems and PGLib-OPF cases. These results are provided for completeness to illustrate behavior across increasing problem sizes.

\begin{figure}[t]
    \centering
    \includegraphics[width=\linewidth]{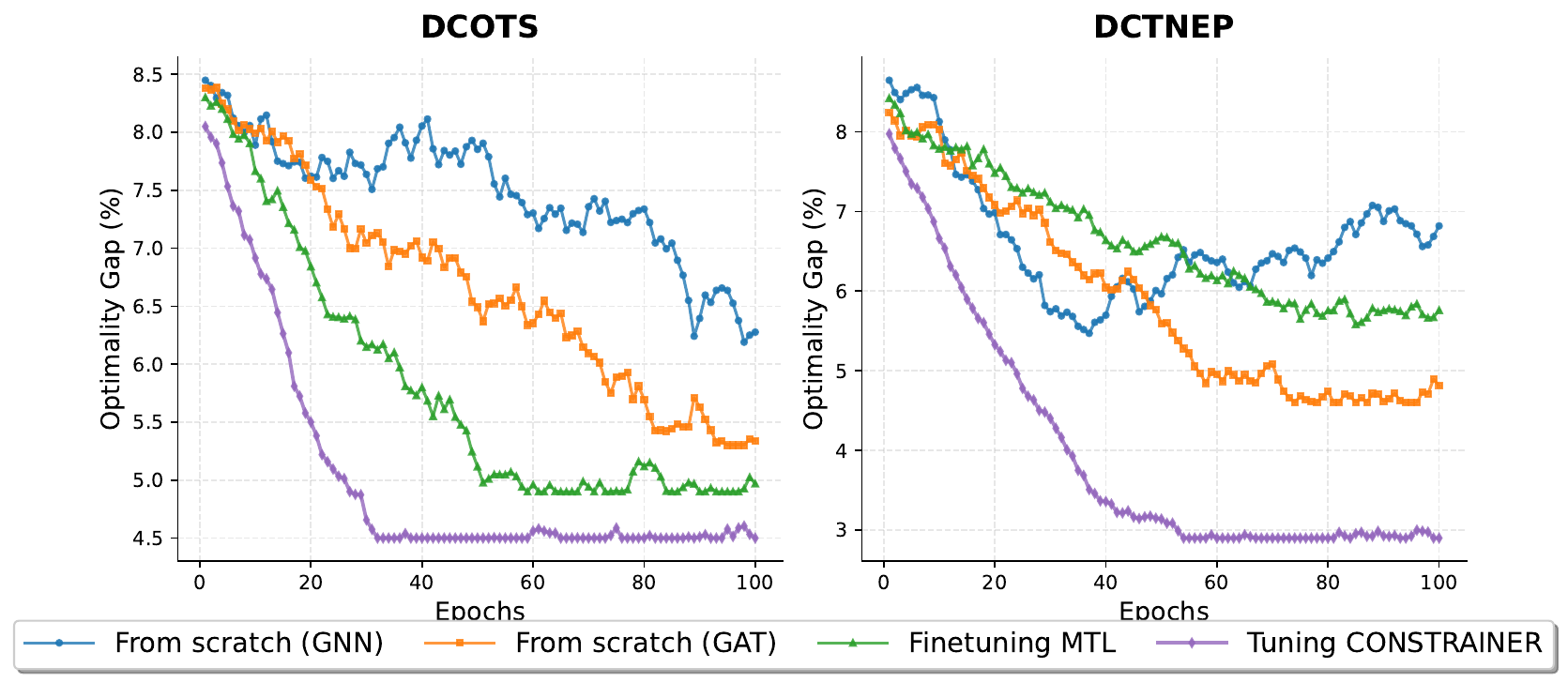}
    \caption{Finetuning on DCOTS and DCTNEP.}
    \label{fig:DCOTS_finetune}
\end{figure}

\paragraph{Additional finetuning curves.}
Figure~\ref{fig:DCOTS_finetune} reports finetuning curves on DCOTS and DCTNEP tasks. The figure is included to complement the main results with representative convergence trends under finetuning.

\begin{table}[t]
\centering
\setlength{\tabcolsep}{4pt}
\renewcommand{\arraystretch}{1.1}
\caption{Generalization performance of CONSTRAINER on OPF problems across increasing grid sizes, evaluated on IEEE benchmark systems (Case118, Case300) and larger PGLib-OPF instances derived from GOC networks (Case500, Case2000).}
\label{tab:grid_generalization}
\begin{tabular}{l|c|c|cc}
\hline
& \multicolumn{2}{c|}{PF}
& \multicolumn{2}{c}{ACOPF} \\
Case
& ACPF Feas $\uparrow$
& DCPF Feas $\uparrow$
& Gap $\downarrow$ & Feas $\uparrow$ \\
\hline

Case118
& \textbf{99.03}
& \textbf{99.50}
& \textbf{3.01} & \textbf{96.10} \\
\hline

Case300
& 98.70
& 99.10
& 3.30 & 95.40 \\
\hline

Case500
& 97.90
& 98.60
& 6.70 & 93.00 \\
\hline

Case2000
& 94.80
& 96.10
& 16.10 & 88.00 \\
\hline

\end{tabular}
\end{table}
\begin{table}[t]
\centering
\setlength{\tabcolsep}{6pt}
\renewcommand{\arraystretch}{1.1}
\caption{ATSP optimality gap (vs.\ LKH-3, lower is better) when training on 100-node instances and evaluating on larger out-of-distribution instances.}
\label{tab:atsp_scale_gen}
\begin{tabular}{l|c|c|c}
\hline
Method & 100 nodes & 500 nodes & 1000 nodes \\
\hline
GOAL        & 0.91\% & 2.37\% & 2.17\% \\
OPTFM       & 0.97\% & 3.19\% & 3.27\% \\
CONSTRAINER & \textbf{0.50\%} & \textbf{1.18\%} & \textbf{1.70\%} \\
\hline
\end{tabular}
\end{table}
\begin{table}[t]
\centering
\small
\setlength{\tabcolsep}{6pt}
\renewcommand{\arraystretch}{1.05}
\caption{ACOPF comparison on Case30 and Case118: feasibility (\%) and optimality gap (\%, lower is better). The hybrid pipeline (CONSTRAINER + FSNet L-BFGS) attains near-FSNet optimality with $\sim$100\% feasibility while preserving CONSTRAINER's cross-task generality.}
\label{tab:hybrid_fsnet}
\begin{tabular}{l|cc|cc}
\hline
& \multicolumn{2}{c|}{Case30}
& \multicolumn{2}{c}{Case118} \\
Method
& Feas $\uparrow$ & Gap $\downarrow$
& Feas $\uparrow$ & Gap $\downarrow$ \\
\hline
FSNet~\citep{nguyen2025fsnet}
& $\sim$100 & 0.82
& $\sim$100 & 0.90 \\
CONSTRAINER (ours)
& 98.20 & 2.05
& 96.10 & 3.01 \\
\rowcolor{blue!8}
CONSTRAINER + FSNet L-BFGS
& \textbf{$\sim$100} & \textbf{1.03}
& \textbf{$\sim$100} & \textbf{1.39} \\
\hline
\end{tabular}
\end{table}
\begin{table}[t]
\centering
\small
\setlength{\tabcolsep}{6pt}
\renewcommand{\arraystretch}{1.1}
\caption{Effect of cross-attention placement on MIS (100 nodes). $L$ is the number of backbone layers.}
\label{tab:crossattn_placement}
\begin{tabular}{l|c}
\hline
Cross-attention placement & Opt.\ gap on MIS \\
\hline
Layer 1       & 0.522\% \\
Layer $L/2$   & \textbf{0.500\%} \\
Layer $L{-}1$ & 0.533\% \\
\hline
\end{tabular}
\end{table}

\paragraph{Generalization to larger ATSP instances.}\label{app:atsp_scale_gen}
We further evaluate cross-size generalization on ATSP, training all methods on 100-node instances and evaluating on 500- and 1000-node instances (Table~\ref{tab:atsp_scale_gen}). CONSTRAINER maintains the lowest gap across all sizes, including the largest out-of-distribution instances, indicating that constraint-conditioned representations transfer beyond the training graph size.

\paragraph{Hybrid feasibility-seeking refinement.}\label{app:hybrid}
When strict feasibility is required on ACOPF, CONSTRAINER's prediction can be refined with a downstream feasibility-seeking step. Table~\ref{tab:hybrid_fsnet} reports a comparison on Case30 and Case118 between FSNet~\citep{nguyen2025fsnet}, CONSTRAINER alone, and CONSTRAINER composed with FSNet's L-BFGS refinement step. The hybrid attains $\sim$100\% feasibility while approaching FSNet's optimality, demonstrating that constraint-conditioned generation integrates cleanly with task-specific repair when hard feasibility is required.

\paragraph{Cross-attention placement ablation.}\label{app:crossattn_placement}
We ablate the layer at which constraint cross-attention is injected into the backbone (Table~\ref{tab:crossattn_placement}, MIS, 100 nodes). Injecting at $L/2$ achieves the best final gap: injecting too early (Layer 1) gives the backbone insufficient structural context before constraint conditioning, while injecting too late (Layer $L{-}1$) leaves too few layers to propagate constraint-conditioned representations.

\section{Additional Analyses}\label{app:review_analyses}
This section reports significance tests, a comparison with an exact solver at matched wall-clock time, the sensitivity of constraint embeddings to perturbations and reformulations, and the size of the constraint specifications per task.

\paragraph{Statistical significance.}
Table~\ref{tab:wilcoxon} compares CONSTRAINER with the strongest generalist foundation-model baseline on each task's test instances using a paired Wilcoxon signed-rank test. The win-rate is the share of instances on which CONSTRAINER attains the lower optimality gap. CONSTRAINER improves significantly on 6 of 8 tasks ($p<0.05$). On MVC and KP, where all methods are close to optimal, the differences are not significant.

\begin{table}[H]
\centering
\small
\caption{Paired Wilcoxon signed-rank test of CONSTRAINER against the strongest generalist foundation-model baseline.}
\label{tab:wilcoxon}
\begin{tabular}{lcccccccc}
\toprule
 & ATSP & CVRP & CVRPTW & OP & JSSP & UMSP & MVC & KP \\
\midrule
$p$-value & 1.3e-3 & 1.5e-3 & 4.2e-8 & 1.3e-2 & 2.0e-4 & 8.7e-10 & -- & -- \\
Win-rate (\%) & 78 & 83 & 95 & 77 & 88 & 86 & 92 & 80 \\
Outcome & sig. & sig. & sig. & sig. & sig. & sig. & parity & parity \\
\bottomrule
\end{tabular}
\end{table}

\paragraph{Comparison with an exact solver at matched time.}
Table~\ref{tab:cpsat} compares CONSTRAINER with CP-SAT on MVC with 100 nodes; optimality gaps are measured against the exact optimum. At the model's 53\,ms budget, CP-SAT reaches a 3.23\% gap, needs about $13\times$ the model's time to match its quality, and proves optimality after 6.1\,s. In throughput, CONSTRAINER produces 18.7 instances/s on one GPU against 3.9 instances/s for CP-SAT on 24 cores. On JSSP ($10\times10$), CP-SAT needs about $1.5\times$ the model's time to match its quality and $3.6\times$ to prove optimality.

\begin{table}[H]
\centering
\small
\caption{CONSTRAINER versus CP-SAT on MVC (100 nodes).}
\label{tab:cpsat}
\begin{tabular}{lcc}
\toprule
Method & Gap (\%) & Time / instance \\
\midrule
CONSTRAINER & 0.32 (median 0) & 53\,ms \\
CP-SAT, matched time & 3.23 & 53\,ms \\
CP-SAT, to reach CONSTRAINER's quality & 0.32 & 707\,ms ($13\times$) \\
CP-SAT, to prove optimality & 0.00 & 6.1\,s ($114\times$) \\
\bottomrule
\end{tabular}
\end{table}

\paragraph{Sensitivity of constraint embeddings.}
Table~\ref{tab:sensitivity} reports the normalized distance $d$ between the embedding of a constraint and that of its perturbed or rewritten version, where $d\approx1$ between two unrelated constraints. Node-feature perturbations keep the graph topology fixed ($\delta=0$) and are averaged over 90 base constraints. Numeric perturbations produce small shifts that grow smoothly with their magnitude, while semantic changes such as an operator flip move the embedding much further. Rewrites that the parser canonicalizes map to identical embeddings, and held-out reformulations produce bounded shifts well below that of a feasibility change.

\begin{table}[H]
\centering
\small
\caption{Normalized embedding distance $d$ under perturbations and reformulations of a constraint.}
\label{tab:sensitivity}
\begin{tabular}{llc}
\toprule
Type & Change & $d$ \\
\midrule
Numeric & coefficient / constant $\pm1\%$ & 0.055 \\
 & coefficient / constant $\pm10\%$ & 0.069 \\
 & coefficient / constant $\pm20\%$ & 0.084 \\
 & right-hand side $\pm20\%$ & 0.143 \\
\midrule
Reformulation & canonicalized (rescale, reorder, term move, slack, dummy variable) & 0.000 \\
 & big-M reformulation (held-out) & 0.007 \\
 & affine recoding (held-out) & 0.20 \\
\midrule
Semantic & variable type (integer $\to$ continuous) & 0.265 \\
 & operator flip ($\le\ \to\ \ge$) & 0.85 \\
\midrule
Reference & unrelated constraint & 1.02 \\
\bottomrule
\end{tabular}
\end{table}

\paragraph{Constraint specifications per task.}
Constraint specifications are defined at the template level: each constraint in a task's constraint set is encoded once as an operator DAG that captures its structural form, and instance-specific values (variable scopes, coefficients, right-hand sides) enter through the problem-graph features. The constraint embeddings are therefore cached and reused across instances. Table~\ref{tab:specs} reports the number of constraints $|\mathcal{C}|$, the total number of DAG nodes, and the encoding time per task.

\begin{table}[H]
\centering
\small
\caption{Constraint specifications per task. All embeddings are encoded once and cached.}
\label{tab:specs}
\begin{tabular}{lcccccccc}
\toprule
 & ATSP & CVRP & CVRPTW & OP & KP & MVC & UMSP & JSSP \\
\midrule
$|\mathcal{C}|$ & 3 & 4 & 7 & 4 & 1 & 3 & 3 & 4 \\
DAG nodes & 21 & 42 & 55 & 39 & 18 & 15 & 21 & 28 \\
Encode time (ms) & 22.0 & 21.2 & 20.4 & 14.2 & 20.1 & 23.1 & 22.8 & 22.8 \\
\bottomrule
\end{tabular}
\end{table}

\paragraph{Constraint conditioning and the decoder mask.}
With the mask removed on KP, CONSTRAINER produces feasible solutions on 75.1\% of instances, against about 55\% for the same model without constraint conditioning.

\end{document}